\documentclass{article}

\usepackage[table,xcdraw]{xcolor}

\usepackage{microtype}
\usepackage{graphicx}
\usepackage{subfigure}
\usepackage{booktabs} 

\usepackage{hyperref}

\usepackage[accepted]{icml2025}

\usepackage{amsmath}
\usepackage{amssymb}
\usepackage{mathtools}
\usepackage{amsthm}
\usepackage{makeidx}
\usepackage{multirow}

\usepackage[capitalize,noabbrev]{cleveref}

\theoremstyle{plain}

\theoremstyle{definition}

\theoremstyle{remark}

\usepackage[textsize=tiny]{todonotes}

\icmltitlerunning{Let Human Sketches Help: Empowering Challenging Image Segmentation Task
with Freehand Sketches}

\begin{document}

\twocolumn[
\icmltitle{Let Human Sketches Help: Empowering Challenging Image Segmentation Task with Freehand Sketches}





\begin{icmlauthorlist}

\icmlauthor{Ying Zang}{yyy}
\icmlauthor{Runlong Cao}{yyy}
\icmlauthor{Jianqi Zhang}{yyy}
\icmlauthor{Yidong Han}{yyy}
\icmlauthor{Ziyue Cao}{comp4}
\icmlauthor{Wenjun Hu}{yyy} \\
\icmlauthor{Didi Zhu}{comp} 
\icmlauthor{Zejian Li}{comp1}
\icmlauthor{Lanyun Zhu}{comp0}
\icmlauthor{Deyi Ji}{comp2}
\icmlauthor{Tianrun Chen}{comp,comp3} 
\end{icmlauthorlist}

\icmlaffiliation{yyy}{School of Information Engineering, Huzhou University}
\icmlaffiliation{comp}{College of Computer Science and Technology, Zhejiang University}
\icmlaffiliation{comp1}{School of Software Technology, Zhejiang University}
\icmlaffiliation{comp0}{Information Systems Technology and Design Pillar, Singapore University of Technology and Design}
\icmlaffiliation{comp2}{School of Information Science and Technology, University of Science and
Technology of China.}
\icmlaffiliation{comp3}{KOKONI, Moxin (Huzhou) Tech. Co., LTD.}
\icmlaffiliation{comp4}{Institute of Psychology, Chinese Academy of Sciences}

\icmlcorrespondingauthor{Tianrun Chen}{tianrun.chen@kokoni3d.com
}

\icmlkeywords{Machine Learning, ICML}

\vskip 0.3in
]



\printAffiliationsAndNotice{}  
\vskip -0.3in
\begin{abstract}
Sketches, with their expressive potential, allow humans to convey the essence of an object through even a rough contour. For the first time, we harness this expressive potential to improve segmentation performance in challenging tasks like camouflaged object detection (COD). Our approach introduces an innovative sketch-guided interactive segmentation framework, allowing users to intuitively annotate objects with freehand sketches (drawing a rough contour of the object) instead of the traditional bounding boxes or points used in classic interactive segmentation models like SAM. We demonstrate that sketch input can significantly improve performance in existing iterative segmentation methods, outperforming text or bounding box annotations. Additionally, we introduce key modifications to network architectures and a novel sketch augmentation technique to fully harness the power of sketch input and further boost segmentation accuracy. Remarkably, our model’s output can be directly used to train other neural networks, achieving results comparable to pixel-by-pixel annotations—while reducing annotation time by up to 120 times, which shows great potential in democratizing the annotation process and enabling model training with less reliance on resource-intensive, laborious pixel-level annotations. We also present KOSCamo+, the first freehand sketch dataset for camouflaged object detection. The dataset, code, and the labeling tool will be open sourced.
\end{abstract}
\vskip -0.3in
{\begin{figure}[!ht]
    \centering
    \includegraphics[width=1.0\linewidth]{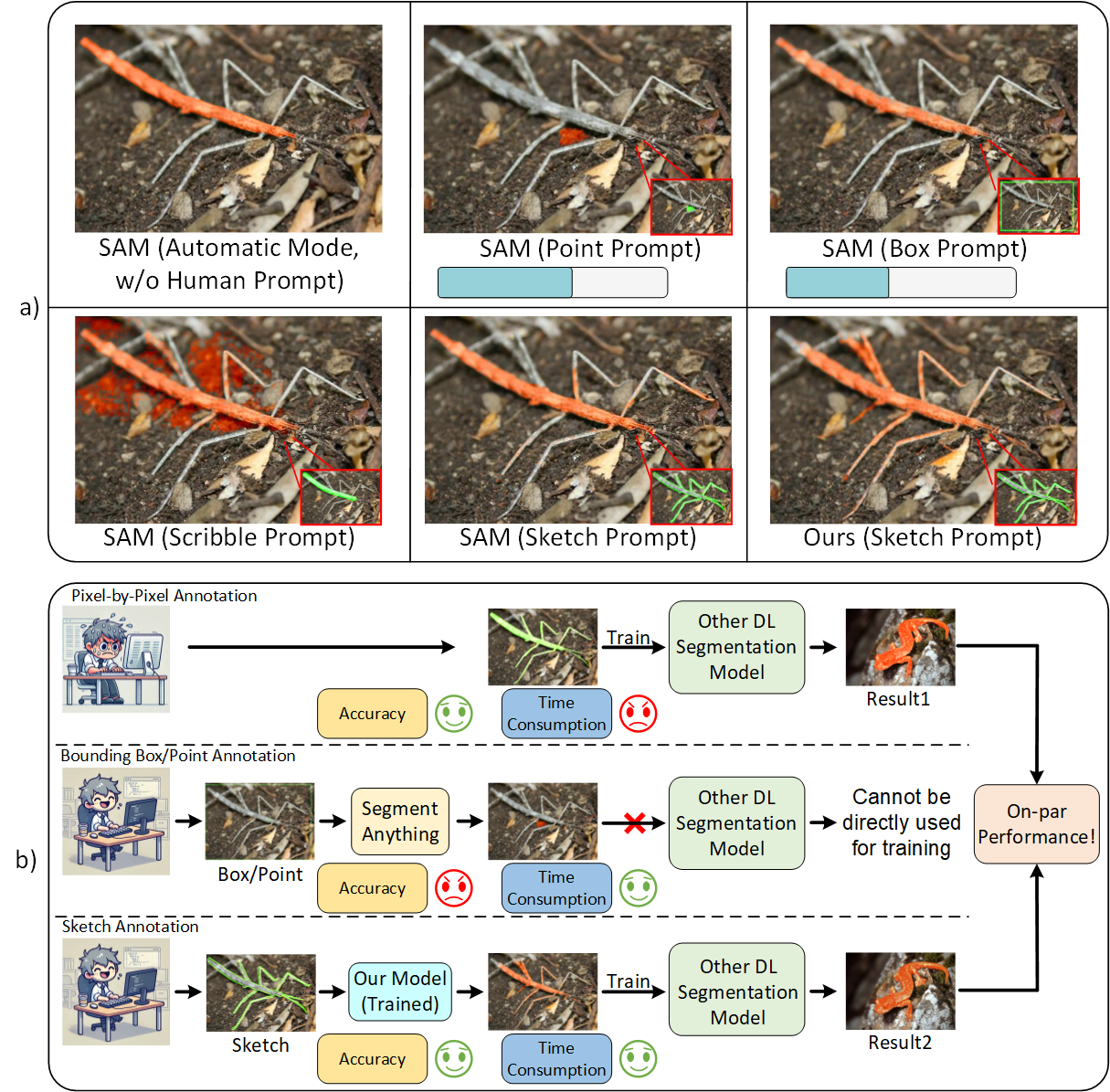}
    \vskip -0.1in
    \caption{In this paper, we propose using human freehand sketches to improve image segmentation in challenging scenes—camouflaged object detection. (a) Comparison of SAM and our method with different inputs. While SAM struggles with point, box, and scribble inputs, sketch input (drawing a rough contour of the object) improves performance with extra annotation time. The blue area shows user response time, and green lines represent prompt types. (b) Practical applications of our method. Our method as an alternative to pixel-by-pixel annotation, producing comparable results, whereas SAM struggles with accuracy using point or box prompts.}
    \vskip -0.2in
    \label{fig:sketch-time}
  \end{figure}

\section{Introduction}

\label{submission}
Sketches have long been a fundamental means of communication, dating back to prehistoric cave paintings \cite{kennedy1974psychology}. A sketch typically refers to a rough contour, yet for humans, even such an approximation is often enough to capture the essence of an object \cite{sayim2011line,1,goodwin2007isophote}. In this paper, we explore how this inherent human ability can help computers overcome a particularly challenging problem in computer vision: detecting objects that vanish into their surroundings—camouflaged object detection, a task that has long eluded even the state-of-the-art foundation models \cite{sam,chen2024sam2,sam-adapter,ravi2024sam}.




Camouflaged object detection (COD) is a critical task, however, with far-reaching applications in fields like military surveillance, wildlife monitoring, and autonomous driving. However, COD is exceptionally challenging due to several factors. Camouflaged objects can appear anywhere in the scene, often blending seamlessly with their surroundings, making them nearly impossible to detect. Additionally, the lack of large, well-labeled datasets further hinders the development of robust models that can generalize across diverse environments. Even powerful foundation models for image segmentation, such as SAM \cite{sam} and SAM2 \cite{ravi2024sam}, trained on vast datasets, continue to struggle with this task, as shown in Figure \ref{fig:sketch-time}.

Despite these challenges, these interactive segmentation models have provided valuable insights, particularly in the use of human inputs—such as points or scribbles—during inference to improve segmentation results \cite{grady2006random,new27,zou2024segment}. Building on this, we demonstrate how human freehand sketches—incorporating knowledge through rough contours—can help the segmentation model achieve satisfactory performance, especially with our newly designed network DeepSketchCamo, thanks to the greater expressiveness of sketches!

A key concern with using sketches to replace point or box prompt is whether they add more time and effort. However, our pilot study shows that in camouflaged object detection, annotators often spend significant time locating the object before even starting to mark points or scribbles (Figure \ref{fig:sketch-time}, details in Supplementary Material). Additionally, switching between tools for foreground and background labeling further complicates the process.

In contrast, our approach is much simpler—asking users to quickly sketch a freehand outline of the object, and the obtained sketch should be able to guide the network better locate the object and produce a more accurate segmentation mask. Users no longer need to switch tools to annotate the foreground and background, nor do they must meticulously trace the object's edges. A rough shape, even just a simple circle around the object, is enough to allow the network to have a better understanding of semantic information. Of course, our experiments confirm that the more precise the sketch, the better the segmentation results (which makes intuitive sense). As there are no existing sketch-GT pairs in the field of camouflaged object detection, we gathered volunteers to annotate thousands of camouflaged objects. This led to the creation of the first dataset of human-drawn sketches that \textit{reflect human attention} \cite{bhunia2023sketch2saliency} in camouflaged objects, which we named KOSCamo+. This dataset not only establishes a foundation for this study, but can also facilitate further research into human attention in complex visual tasks or computer-human collaboration. 

Our experiments show that simply modifying the network to replace SAM’s input from boxes/points to sketches already improves segmentation performance. But we didn't stop there. We further optimize the process to make the extra human annotation time truly worthwhile, enabling our newly designed network, we named DeepSketchCamo, to produce results on par with pixel-by-pixel ground truth annotations.


Our first goal was to establish a robust segmentation backbone. We leveraged the cutting-edge SAM model and designed an adaptation mechanism to introduce domain-specific information to ensure it could handle the demanding camouflaged detection task with precision. Building on this strong foundation, we designed to use a sketch encoder to encode the freehand sketches, and the encoded features were then seamlessly fused with image features through a fusion mechanism, before being decoded by the SAM decoder. The decoder uses the fused feature and the (encoded) sketch prompt to generate the final segmentation masks. Moreover, to handle the varying precision and abstraction of sketches, we designed a sketch enhancement mechanism to improve robustness during training, making sure the network focused on using the \textit{saliency} \cite{bhunia2023sketch2saliency} from the sketch. Additionally, to address the imprecision of sketches at object boundaries, we developed a boundary refinement module and added an adaptive focal loss component, providing stronger supervision for improved segmentation performance.

Through extensive experimentation, we demonstrate the value of our research. First, we demonstrate that freehand sketches deliver a significant performance boost in existing models like SAM, outperforming other prompting techniques like points, boxes, and scribbles. Furthermore, our innovative methods tailored for sketch inputs yielded an even greater enhancement, with our network DeepSketchCamo achieving 18\% better performance compared to SAM-based networks. More importantly, we demonstrated that the result obtained from our network can be directly utilized in training other networks. The network trained on our data matches the quality of that used ground truth masks to train, while saving up to 120 times the annotation time. 

In summary, our contributions are: (i) For the first time, we harness the expressiveness of human sketches for challenging object segmentation task camouflaged object detection; (ii) We present a new highly accurate segmentation and annotation method DeepSketchCamo that can replace traditional, time-consuming labeling techniques and achieve performance on-par with pixel-level labeling; (iii) We introduce KOSCamo+, the first freehand sketch dataset for camouflaged object detection.

Just as SAM has revolutionized annotation with its simplicity and accuracy, we believe our research paves the way for new possibilities by introducing sketch-based annotations with high-precision results, and thus can reducing reliance on resource-intensive, laborious pixel-level annotations, opening up more efficient and accessible avenues for training future segmentation models.

\vskip -0.2in
\section{Related Work}
\noindent\textbf{Sketch for Visual Understanding.}
Sketches showed advantages in various computer vision tasks \cite{bhunia2020pixelor,zhang2020interactive}. For example, in image/video synthesis \cite{zhang2020interactive,koley2024s}, and 3D generation \cite{deep3dsketch,deep3dsketch+,Deep3DVRSketch, deep3dsketch-im,chen2023reality3dsketch,chen2024img2cad,zang2023deep3dsketch+,hu2024sketch,chen2024high,chen2024new,chen2024rapid}, sketches are used to guide network to produce user-desired content. In the realm of content retrieval \cite{,visualimage,sketch-based,sketch-basedmanged,8}, which has been researched for decades. Similarly, sketch can be used for 
object localization \cite{19, 20}. Recently, sketches have also been introduced to (zero-shot) object detection \cite{whathumancan}. 

\begin{table}[ht]
\vskip -0.25in
\caption{Comparison between our approach and some similar sketch-enabled and interactive segmentation approach}
\centering
\renewcommand{\arraystretch}{1.6} 
\resizebox{\columnwidth}{!}{ 
\begin{tabular}{|c|c|c|c|c|}
\hline
\multicolumn{1}{|c|}{Method} & \parbox{3cm}{Sketch-Assisted Saliency Detection \cite{bhunia2023sketch2saliency}} & \parbox{3cm}{Sketch-based Segmentation \cite{hu2020sketch}} & \parbox{3cm}{Interactive Segmentation Foundation Model \cite{kirillov2023segment}} & Ours \\ \hline
Spatial Specification &  &  & \checkmark & \checkmark \\ \hline
Human-guided Improvement & \checkmark & \checkmark & \checkmark & \checkmark \\ \hline
Input Error-Tolerant Flexibility  &  & \checkmark &  & \checkmark  \\ \hline
High-quality Annotation &  &  &  & \checkmark \\ \hline
\end{tabular}
}
\vskip -0.15in
\label{Comparative}
\end{table}

\noindent In this study, we focus on a fundamental computer vision task: image segmentation. Specifically, we aim to utilize sketches as a means of human-computer interaction and collaboration \cite{humansketch}, allowing human knowledge to be infused into the network to enhance segmentation performance (with \textbf{human-guided improvement} on performance), especially the challenging camouflaged object detection field. Unlike previous approaches that treated sketches as mere input prompts—essentially substituting text or fixed class labels \cite{hu2020sketch}—we encourage users to draw a sketch in the area they wish to segment (with \textbf{spatial specification}). This sketch serves to assist the network in achieving better localization and obtaining more accurate results (functioning more like an enhanced version of scribbles in interactive segmentation, but with more details and with some \textbf{input error-tolerance flexibility}—user do not need to carefully trace the edge, rough sketches within or outside of the object are both fine). Note that freehand sketches are not edge maps, they are inherently less precise but are easier to obtain. The comparison between our method and some key methods is shown in Table. \ref{Comparative}, only our method can achieve a performance that can be directly used to annotate and train other models as we show in the later Experiment Section.

\vskip -0.1in
\noindent\textbf{Interactive Segmentation.}
Interactive segmentation is proposed to enhance the network performance of image segmentation \cite{chen2017deeplab, zhu2021learning, zhu2023continual, zhu2024addressing} with the help of human input. The types of user input include points, boxes, and scribbles (no freehand sketches so far) \cite{new27,new7,new25,new26,new45}. Recently, one of the most influential interactive segmentation models is SAM \cite{kirillov2023segment}, which is trained on a vast amount of data and is capable of segmenting objects based on points or box input with impressive accuracy. However, we show that in this paper freehand sketches can be used as a powerful modality to further improve the effectiveness of SAM.

\vskip -0.1in
\noindent\textbf{Referring Expression Segmentation.}
Referring Expression Segmentation (RES) takes the user input and generates a user-desired segmentation result \cite{3-1}. Most of the work use input of text prompt \cite{uniref++121,uniref++35,uniref++99,refcod,zang2025resmatch,chen2024reasoning3d}, few use images \cite{3-4,3-3,3-2}. 
Our approach differs from traditional RES methods but can be regarded as a specialized RES task that uses sketches as input. It is important to note that, unlike some past RES methods that suffer from sparse annotations and ambiguous text, leading to performance that lags behind general segmentation models, the sketches we employ have been widely demonstrated to encode more information than images across various tasks \cite{sangkloy2022sketch,song2017fine,magic3dsketch,olsen2009sketch}. By drawing sketches at corresponding locations, our method is designed to boost the network's segmentation performance.

\vskip -0.1in
\noindent\textbf{Underperformed Scenes for Foundation Model-Camouflaged Object Detection.}
Although foundation models have demonstrated remarkable performance across various downstream tasks, they still face limitations in certain challenging scenarios \cite{sam-adapter,chen2024sam2}. A notable example is camouflaged object detection, where models like SAM \cite{sam-adapter} often struggle. camouflaged object detection involves identifying objects that blend seamlessly into their surroundings, which is crucial for enhancing situational awareness for computer vision systems. Historically, various methods have explored different avenues to improve camouflage detection \cite{4-18,sam-adapter42,sinet,pfnet}, yet there remains a need for further advancements. A key question is whether we can enhance segmentation performance by incorporating human knowledge input (not with this weakly supervised method \cite{he2023weakly,chen2025hint,he2024weakly}, but \textit{with added human input in a fully-supervised setting}), we tried to answer this question in this paper. We also note that recent developments have introduced open vocabulary approaches for camouflage object segmentation \cite{pang2023open}. While our method can also be considered an open vocabulary setting—supporting new classes based on sketches without restrictions on fixed categories—our main goal is to enhance the accuracy of the segmenter. We achieve this goal by utilizing sketches while leveraging the strengths of the foundation model.

\vskip -0.3in
\section{Method}
\subsection{Drawing Freehand Sketch -- A Pilot Study and a New KOSCamo+ Dataset}

\begin{figure*}[!h]
	\centering
	\includegraphics[width=\textwidth]{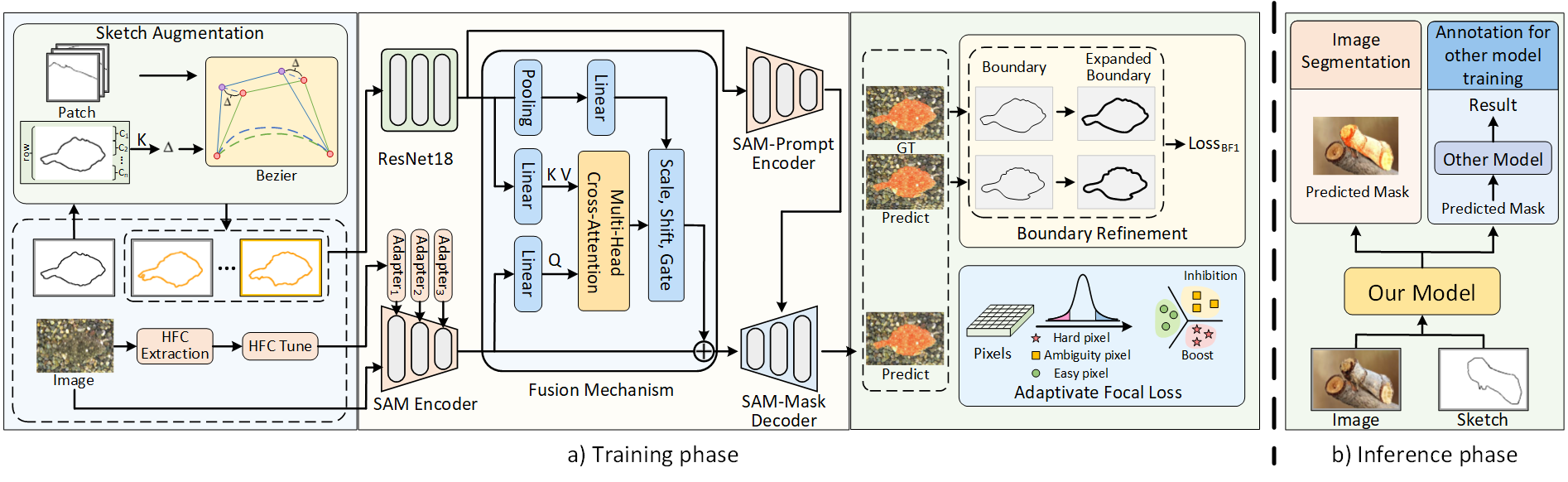}
    \vspace{-0.4cm}
    \vskip -0.1in
	\caption{(a) Overall structure of DeepSketchCamo. Sketch augmentation, boundary refinement, and adaptive focal loss are introduced in this sketch-based task and are used in training to get elevated performance. The GT mask is used for supervision. (b) The application of our model. After the model has been trained, we input an image and a rough sketch into the model, the model can generate a predicted mask which can be directly used to train other deep-learning models.} 
    \vskip -0.1in
	\label{fig:2}
    \vspace{-0.2cm}
\end{figure*}

Before diving into our main research, we explored annotation methods for COD. We aim to answer a question: what type of input is most suitable for annotating samples in challenging segmentation tasks like COD?

Here, a unique aspect of the COD task is that distinguishing camouflaged objects is tough. This challenge differs significantly from general segmentation due to how intricately objects blend into their surroundings. We conducted a pilot study testing various annotation methods: points, boxes, and scribbles. During the average annotation time (see Figure \ref{fig:sketch-time}), a significant portion (65\% for point, 46\% for box) was spent determining which object to annotate.

We also interviewed annotators, who noted that switching between foreground and background tools consumed valuable time. While bounding boxes simplify this distinction, they must fit snugly around the object. If the box strays too far from the edges—especially with tricky tail-like forms—it can hinder performance and introduce unwanted segments.

Given these challenges, we propose that instead of aligning bounding box edges, annotators use a freehand sketch (like ``high-resolution box"). But the annotator doesn’t need to fit the sketch with the actual contours—whether inside or outside the object's edges are fine. (Otherwise, it would have no practical value because it would waste too much time.) Visualization of comparison between our sketch input and other forms is shown in the supplementary material.

In practice, we used custom software on Pads for this task, allowing volunteers to sketch with pencils directly on the images without any ground truth (GT) mask reference. To keep the sketches efficient and practical, we set a drawing limit of 30 seconds and timed each session, with the average completion time being 21 seconds. Ultimately, we created sketches and scribbles for 2,352 images individually in the test sets of existing camouflage object segmentation datasets CHAMELEON \cite{chamelon}, CAMO \cite{sam-adapter27}, and COD10K \cite{sinet}. Additionally, we drew sketches for 4,121 images in the NC4K dataset \cite{4-29}, resulting in a total of 8,825 pairs of images and sketches, which we named the KOSCamo+ dataset. More details are in the supplementary material.

\subsection{Network Backbone}
With our dataset in hand, we then design the network, shown in Figure \ref{fig:2}. The network is built upon the SAM \cite{sam} network, which consists of an image encoder, a prompt encoder, and a decoder. Images $I \in \mathbb{R}^{H\times W\times3}$ are encoded by the encoder to obtain feature $F_I$ and then passed into the decoder, while the input prompts are encoded by the prompt encoder and integrated into the decoding process. Intuitively, by replacing the prompt encoder with a sketch encoder, we could use sketches $S \in \mathbb{R}^{H\times W}$ as the prompt. Our experiments confirmed that this naive approach of using sketches as the input can already boost the model performance as shown in Table \ref{table.main}.

However, we aimed to further enhance the control provided by the sketches. To achieve this, we introduced a Fusion Mechanism before the decoding step, which has been validated to be effective in previous works \cite{uniref++,uniref}. The sketches, encoded using ResNet \cite{resnet} and obtained feature $F_S$, were directed into the SAM decoder while also going through this Fusion Mechanism, which allowed for constraints on the feature inputs prior to entering the decoder.

Specifically, in the Fusion Mechanism, the query values are derived from the image features $F_I$, while the key and value embeddings are from the sketch features $F_S$. These inputs are first linearly projected in three vectors: $Q$, $K$, and $V$. We perform a multi-head cross-attention on them after the reference features are pooled and used to regress the scale, shift, and gate parameters $\gamma$, $\beta$, $\alpha$, respectively. Finally, the output features are added to the original visual features $F_I$ via a residual connection. The process is represented as:
\begin{equation}
     O = \text{Attention}(Q, K, V)
\end{equation}
\vspace{-0.5cm}
\begin{equation}
     {\gamma, \beta, \alpha} = \text{Linear(Pooling}(F_S))
\end{equation}
\begin{equation}
     F_U = F_I + \alpha(O(1+\gamma) + \beta)
\end{equation}
\noindent where $O$ is the results of attention operation, $F_U$ is the output of the fusion module. Next, we use the SAM-Prompt Encoder to extract features $F_S$, represented as $F’_S$.
Finally, the SAM-Mask Decoder uses $(F_U + F'_S)$ to produce a binary mask for the target object $M \in \mathbb{R}^{H\times W}$.

\subsection{Injecting Domain Specific Information}
Next, we focused on designing targeted improvements for the network to inject human knowledge, addressing two key challenges: how to fully exploit sample-specific sketches and how to alleviate the inherent difficulties of the task. 

To tackle the challenges of inherent difficulties of the task, we first introduced general priors that do not require additional human annotations. Specifically, we utilized explicit high-frequency information extraction and patch embedding that have proven effective in past work on COD \cite{sam-adapter,evp,chen2024sam2}. We implemented this knowledge injection by integrating a series of Adapters into the SAM encoder. Two MLPs and activate functions within two MLPs are used to generate high-frequency feature $F_{hfc}$ and patch embedding feature $F_{hpe}$. These features are fused to obtain $F'_I = {\rm MLP}_{up}({\rm GELU}({\rm MLP}(F_{hfc} + F_{hpe})))$, where \text{MLP} are linear layers used to generate task specific features. $\text{MLP}_{up}$ is an up-projection layer that adjusts the dimensions of transformer features $F'_I$ refers to the output feature that is attached to each transformer layer of the SAM model. GELU is the activation function \cite{hendrycks2016gaussian}.

\subsection{Sketch Augmentation}
Considering that sketches inherently possess abstraction and diversity, with significant skill level differences between creators, even the same individual may produce inconsistent styles and details, we designed a sketch augmentation method to fully leverage these sketch data without letting these differences negatively impact network performance. We want to let the network learn the saliency of the object, which is a useful property that sketches inherently have \cite{bhunia2023sketch2saliency}. This method utilizes a parameterization technique based on Bezier curves \cite{shaheen2017constrained} to fit the sketches. By adjusting these parameters or introducing random perturbations, we can programmatically synthesize a variety of sketches \cite{zheng2021sketch}, thus enhancing the network's robustness to different inputs and encouraging it to focus on the key information necessary for segmentation, rather than being influenced by stylistic variations. This sketch augmentation method allows us to generate imprecise sketches near the mask edges, closely resembling human drawing behavior. This approach helps reduce the domain gap and enables us to more effectively utilize human sketches during inference. In Figure \ref{fig:3}, we visualize the sketch being generated with different deformation strengths. 

{\begin{figure}[h]
	\centering
	\includegraphics[width=0.9\linewidth]{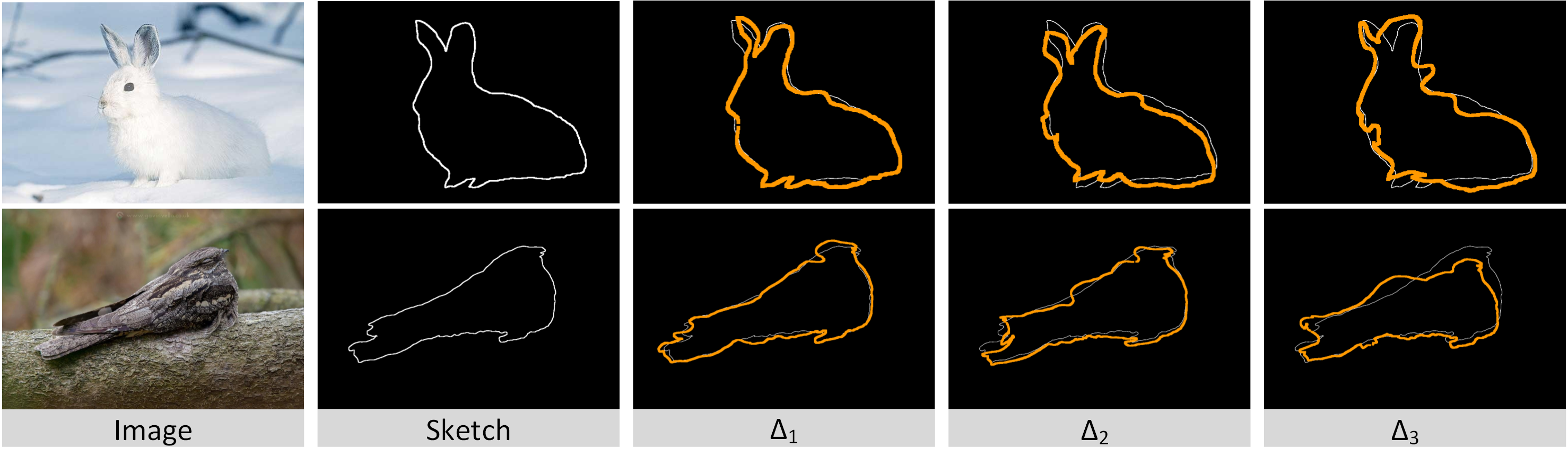}
    \vskip -0.1in
	\caption{Visualization for sketches augmentation. Sketches generated by different amplitudes of variation of the Bezier curve. }
	\label{fig:3}
\end{figure}
\vskip -0.1in

Specifically, we extract the centerline $S_0$ with a deformation-based skeletonization method. It removes pixels on the binary sketch image boundary while preserving the Euler number. We segment $S_0$ into $n$ disjoint rectangular patches. In each patch, we select the largest set of connected pixels as the principal curve $T$, which is fitted by a cubic Bezier curve \cite{shaheen2017constrained}:
\begin{equation}
\small
f = (1-t)^3p_0 + 3t(1-t)^2p_1 + 3t^2(1-t)p_2 + t^3p_3
\end{equation}

\noindent where $t \in [0,1]$ , $p_0$ and $p_3$ are the start and end points of the curve. Pivots $p_1$ and $p_2$ are control points, which determine the shape of the curve.

To ensure fitting performance, curves containing very few pixels are eliminated. We apply the least squares method to determine the best-fitting Bezier curve using the objective function:
\begin{equation}
f^* = \min L = \min \sum_{i=1}^n (v_i - f(t_i))^2
\end{equation}

To perform ``data enhancement" (generate new sketches based on existing ones), we apply random displacement $\Delta$ to the control points, and generate new control point positions $p$ by adjusting the original control points $p'= p + \Delta$.

To effectively control the magnitude of the displacement $\Delta$, we can use the number of lines in the sketch:
\begin{equation}
    \Delta = {\rm floor}\left\lfloor\frac{row}{C}\right\rfloor \times K
\end{equation}
where $C$ is the number of control unit rows of displacement deformation, $K$ is the amount of increase in the corresponding displacement amplitude when the number of sketch rows increases by $C$, and ${\rm floor}(\cdot)$ represents rounding down. As the sketch size increases, the displacement amplitude gradually increases, ensuring that larger sketches experience correspondingly larger displacement deformations. This design ensures that sketches of different sizes can maintain consistent deformation effects, thereby effectively addressing the problem of sketch diversity and achieving a balance between deformation flexibility and stability. 

\subsection{Boundary Refinement and Adaptive Focal Loss}
In practice, sketches cannot provide strong control over boundaries, as they typically indicate only a rough area (if we don't impose constraints on the boundaries, the introduction of sketches could even degrade performance). However, the challenge in COD lies in the inaccuracies of these boundaries. To address this, we creatively introduced a boundary refinement loss to enhance the network's learning of boundary details, providing an additional supervisory signal. To our knowledge, this method is the first of its kind \cite{bokhovkin2019boundary} applied to the camouflage object segmentation. Specifically, a binary map boundary of the prediction and GT is obtained through max-pooling:
\begin{equation}
    M_{gt}^b = \operatorname{Maxpool}((1-M_{gt}), \theta_1) - (1-M_{gt})
    \label{equation.3}
\end{equation}
\begin{equation} 
    M_{pd}^b = \operatorname{Maxpool}((1-M_{pd}), \theta_1) - (1-M_{pd})
    \label{equation.4}
\end{equation}
where $\theta_1$ is the size of the sliding window. To compute Euclidean Distances from pixels to boundaries, a supporting map of the extended boundary should be obtained:
\begin{equation}
    M_{gt}^{b,ext} = \operatorname{Maxpool}(M_{gt}^b, \theta_2)
\end{equation}
\begin{equation}
    M_{pd}^{b,ext} = \operatorname{Maxpool}(M_{pd}^b, \theta_2)
\end{equation}
where $\theta_2$ can be determined as less than the minimum distance between neighboring segments of the binary ground truth map. So the precision and recall can be computed:
\begin{equation}
    P = \frac{\operatorname{sum}(M_{pd}^b \circ M_{gt}^{b,ext})}{\operatorname{sum}(M_{pd}^b)},     R = \frac{\operatorname{sum}(M_{gt}^b \circ M_{pd}^{b,ext})}{\operatorname{sum}(M_{gt}^{b})}
\end{equation}
where operation $\circ$ denotes the pixel-wise multiplication of two binary maps and operation sum(·) is the pixel-wise summation of a binary map. Finally, the reconstructed metric and the boundary refinement function are defined as:
\begin{equation}
    BF_{1} = \frac{2PR}{P+R}, \operatorname{Loss}_{BF_1} = 1- BF_{1}
\end{equation}
Additionally, we incorporated Adaptive Focal Loss to address the shortcomings of traditional focal loss  $L_{FL} = -\sum_{i=1}^{N} (1-P^i_t)^\gamma \log(P^i_t)$, (where $P_n$ is the final prediction, $y_i$ is ground truth, $P^i_t = \begin{cases}
    P_n^i, & \text{if } y_t^i = 1 \\
    1-P_n^i, & \text{else } y_t^i = 0
\end{cases}$, and $P^i_t \in [0,1]$ is difficulty confidence of each pixel). The above-mentioned traditional focal loss can help manage class imbalance by down-weighting easy examples and focusing on hard ones. Further, the addition of sketches influences the network's output confidence and affects the gradient dynamics (details in Supplementary Material). We note that the gradients of ambiguous pixels are overwhelmed by those of hard pixels, which, in turn, hinders the performance in classifying ambiguous pixels. Therefore, we use Adaptive Focal Loss to adapt the learning strategy according to the global training situation, mitigating the gradient swamping issues mentioned in \cite{aml}:
\begin{equation}
    L_{AFL} = \sum_{i=1}^{N} -(1-P^i_t)^{(\gamma+\gamma_a)} \log(P^i_t) + \alpha(1-P^i_t)^{(\gamma+\gamma_a+1)}
\end{equation}
\begin{equation}
    \gamma_a = 1 - \frac{\sum_{i=1}^{\mathcal{H}} (P^i_t)}{\sum_{i=1}^{\mathcal{H}} (y^i_t)}
\end{equation}

\begin{table*}[!ht]
\vskip -0.1in
\caption{Our method achieves state-of-the-art results in COD benchmarks; best results are highlighted in bold.}
\centering
\vskip 0.1in
\resizebox{0.75\textwidth}{!}{
\centering
\begin{tabular}{l|l|llll|llll|llll}
\hline
\multicolumn{1}{c|}{\multirow{2}{*}{Methods}}   & \multirow{2}{*}{Prompt}         & \multicolumn{4}{c|}{CHAMELEON (76 images)}                                              & \multicolumn{4}{c|}{CAMO (250 images)}                                                   & \multicolumn{4}{c}{COD10K (2026 images)}                                                  \\ \cline{3-14} 
\multicolumn{1}{c|}{}                           &                                 & $S_m \uparrow$ & $E_{m} \uparrow$ & $F_\beta^w \uparrow$ & MAE$ \downarrow$ & $S_m \uparrow$ & $E_{m} \uparrow$ & $F_\beta^w \uparrow$ & MAE$ \downarrow$ & $S_m \uparrow$ & $E_{m} \uparrow$ & $F_\beta^w \uparrow$ & MAE$ \downarrow$ \\   \cline{0-13}
UJSC \cite{ujsc}               & \multicolumn{1}{c|}{-}          & 0.891          & 0.955            & 0.833                & 0.030            & 0.800          & 0.859            & 0.728                & 0.073            & 0.809          & 0.884            & 0.684                & 0.035            \\
UGTR \cite{ugtr}               & \multicolumn{1}{c|}{-}          & 0.888          & 0.940            & 0.794                & 0.031            & 0.784          & 0.822            & 0.684                & 0.086            & 0.817          & 0.852            & 0.666                & 0.036            \\
ZoomNet \cite{zoomnet}         & \multicolumn{1}{c|}{-}          & 0.902          & 0.958            & 0.845                & 0.023            & 0.820          & 0.892            & 0.752                & 0.066            & 0.838          & 0.911            & 0.729                & 0.029            \\
FEDER \cite{feder}             & \multicolumn{1}{c|}{-}          & 0.887          & 0.954            & 0.834                & 0.030            & 0.802          & 0.873            & 0.738                & 0.071            & 0.823          & 0.905            & 0.716                & 0.032            \\
FSPNet \cite{fspnet}           & \multicolumn{1}{c|}{-}          & 0.908          & 0.965   & 0.851                & 0.023            & 0.856 & 0.928   & 0.799                & 0.050   & 0.851          & 0.930   & 0.735                & 0.026            \\
FPNet \cite{fpnet}             & \multicolumn{1}{c|}{-}          & \textbf{0.914} & 0.961            & 0.856       & 0.022   & 0.852          & 0.905            & 0.806       & 0.056            & 0.850          & 0.913            & 0.748                & 0.029            \\
EVP \cite{evp}                 & \multicolumn{1}{c|}{-}          & 0.871          & 0.917            & 0.795                & 0.036            & 0.846          & 0.895            & 0.777                & 0.059            & 0.843          & 0.907            & 0.742                & 0.029            \\
SAM-Adapter \cite{sam-adapter} & \multicolumn{1}{c|}{-} & 0.896          & 0.919            & 0.824                & 0.033            & 0.847          & 0.873            & 0.765                & 0.070            & \textbf{0.883} & 0.918            & 0.801       & 0.025   \\ \hline
\multirow{5}{*}{SAM \cite{sam}}            
& \multicolumn{1}{c|}{-}  & 0.796          & 0.802            & 0.676                & 0.062            & 0.750          & 0.756            & 0.639                & 0.105            & 0.789          & 0.817            & 0.596                & 0.049            \\
& Box                             & 0.757          & 0.820            & 0.681                & 0.115            & 0.701          & 0.784            & 0.654                & 0.172            & 0.800          & 0.866            & 0.721                & 0.075            \\
                                           & Point                           & 0.708          & 0.717            & 0.563                & 0.082            & 0.643          & 0.658            & 0.510                & 0.139            & 0.682          & 0.715            & 0.525                & 0.099            \\
                                           & Scribble                        & 0.777          & 0.845            & 0.721                & 0.060            & 0.750          & 0.826            & 0.721                & 0.082            & 0.808          & 0.874            & 0.741                & 0.037            \\
                                           & Sketch                          & 0.882          & 0.942            & 0.841                & 0.026            & 0.822          & 0.909            & 0.786                & 0.061            & 0.850          & 0.935            & 0.775                & 0.026            \\ \hline
\multirow{2}{*}{UniRef++ \cite{uniref++}}                                       
& Scribble                        & 0.880          & 0.933            & 0.844                & 0.027            & 0.817          & 0.894            & 0.775                & 0.066            & 0.851          & 0.925            & 0.784                & 0.026            \\
& Sketch                          & 0.906 & 0.966   & 0.881       & 0.023   & 0.856 & 0.942   & 0.833       & 0.049   & 0.867 & 0.950   & 0.807       & 0.023   \\
                                      \hline
\multirow{2}{*}{Refcod  \cite{refcod}}                                        
 & Scribble                        & 0.786          & 0.878            & 0.661                & 0.056            & 0.624          & 0.704            & 0.434                & 0.135            & 0.730          & 0.822            & 0.549                & 0.053            \\
& Sketch                          & 0.779          & 0.893            & 0.650                & 0.058            & 0.625          & 0.720            & 0.437                & 0.133            & 0.721          & 0.822            & 0.526                & 0.054            \\
                                       \hline
\multirow{5}{*}{Ours}                                            & Point                           & 0.869              & 0.924                & 0.825                    & 0.033       & 0.804     & 0.866       & 0.752                    & 0.073                & 0.830              & 0.895                & 0.749                    & 0.035                \\
                                            & Box                             & 0.900              & 0.956                & 0.875                    & 0.024                & 0.841              & 0.920                & 0.812                    & 0.055                & 0.859              & 0.939                & 0.797                    & 0.023                \\
                                            & Scribble                        & 0.885              & 0.944                & 0.851                    & 0.027                & 0.832              & 0.902                & 0.816                    & 0.060                & 0.854              & 0.929                & 0.796                    & 0.026                \\

                                            & \cellcolor{gray!20}Sketch                          &\cellcolor{gray!20}0.907          & \cellcolor{gray!20}\textbf{0.974}   &\cellcolor{gray!20}\textbf{0.886}                & \cellcolor{gray!20}\textbf{0.022}   &\cellcolor{gray!20}\textbf{0.861} & \cellcolor{gray!20}\textbf{0.945}   & \cellcolor{gray!20}\textbf{0.843}       & \cellcolor{gray!20}\textbf{0.048}   & \cellcolor{gray!20}0.876 & \cellcolor{gray!20}\textbf{0.957}   &\cellcolor{gray!20} \textbf{0.827}       & \cellcolor{gray!20}\textbf{0.021}   \\
                                            
                                            \hline
\end{tabular}}

\label{table.main}
\end{table*}

The foreground is the hard pixels and the number is $\mathcal{H}$ (obtained through $y_t$). To estimate the overall learning difficulty of the model, the $P_t^i$ of each pixel represents its learning level. Then, the learning state of the model can be denoted as $\sum_{i=1}^{\mathcal{H}} (P^i_t)$.
Ideally, the optimization goal of the model is to classify all hard pixels completely and accurately. In this case, the optimal learning situation of the model can be represented as $\sum_{i=1}^{\mathcal{H}} (y^i_t)$.

\vskip -0.2in
\section{Experiments}
\subsection{Implementation Details}
 \vskip -0.05in

\begin{figure*}[ht]
    \centering
    \includegraphics[width=0.7\textwidth]{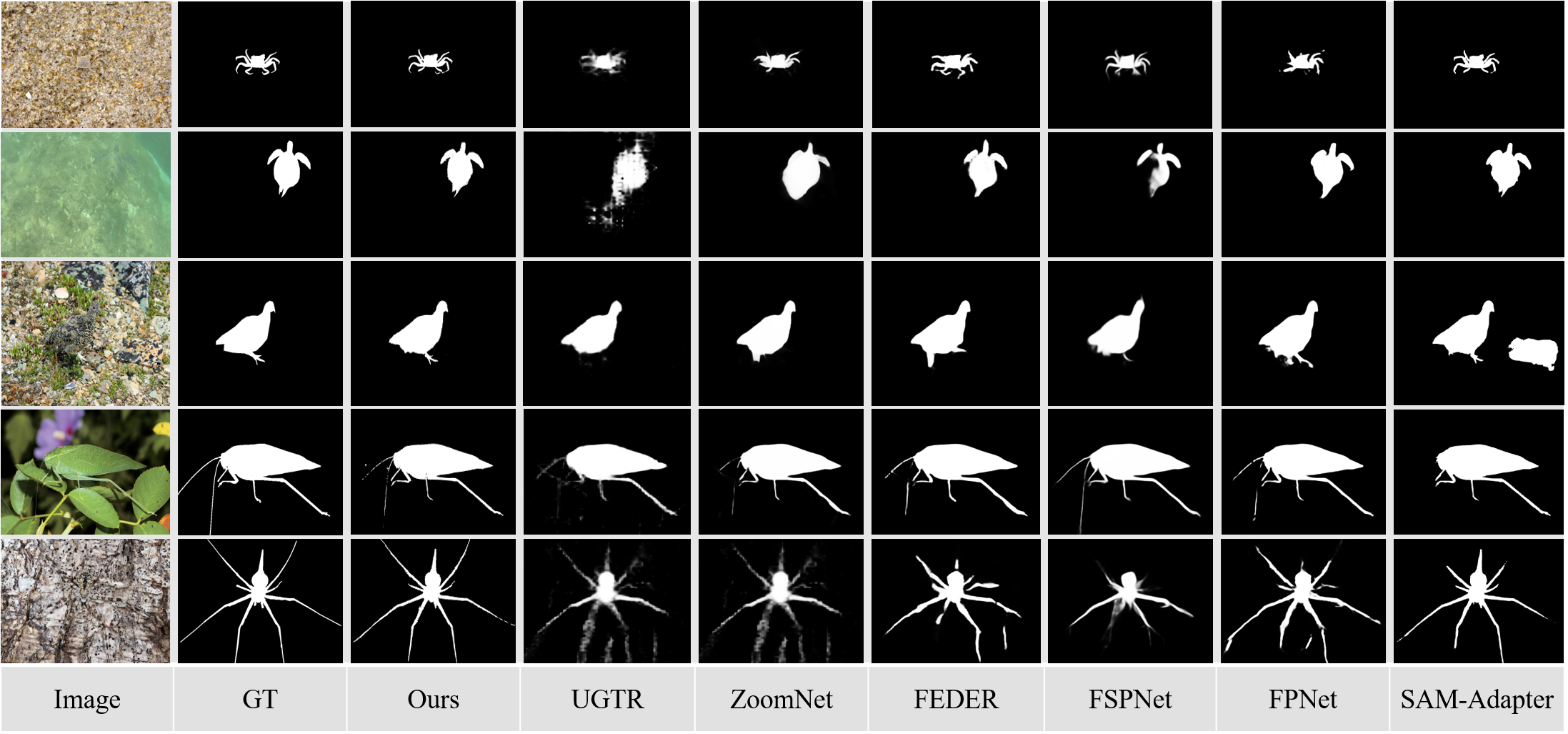}
    \vskip -0.1in
    \caption{The visualization results of camouflaged image segmentation, our approach significantly enhances the performance of object segmentation, such as accurately segmenting spider legs, insect antennae, and the body of a turtle.}
    \vskip -0.1in
    \label{fig:visual iamge}
  \end{figure*}

Training takes ~12 hours for 120,000 iterations. The initial learning rate is set to 2e-4, with a linear warm-up strategy applied for the first 200 iterations. A step decay is applied on the 75,000th iteration. We adopt four evaluation metrics: S-measure ($S_m$), E-measure ($E_m$), weighted F-measure ($F_\beta^w$), and mean absolute error (MAE). The train/test split follows \cite{sam-adapter}. The sketches in the test set are from KOSCamo+ dataset. The scribble is also manually labeled. The box and point input are derived from the sketch in KOSCamo+ dataset, more details of data processing and training protocols are in the Supplementary Material.


\vskip -0.15in
\subsection{Sketch Boost the Segmentation Performance}
\vskip -0.05in
As shown in Table \ref{table.main}, we compared various benchmarks to highlight the superiority of our method. The visualization result is in Figure \ref{fig:visual iamge}. This effectively demonstrates our two claims: 1) sketches are an effective tool for segmentation assistance, which can boost performance to different existing models; and 2) The network proposed by us is effective—our network design and training protocol tailored for sketch input introduces a new state-of-the-art approach in this field.

First, sketch enhances the performance of segmentation networks. When tested on current leading model, SAM, compared to other inputs like box, point and scribble, sketch input group yields the best result. A similar trend can be found on other referring segmentation models inducing UniRef++ and RefCod. This validates our claim that sketches can serve as valuable inputs for image segmentation. Our method surpasses mainstream methods that do not utilize additional prompt inputs (which is a natural outcome). 

Second, the effectiveness of our network is also validated. Our method outperforms existing models across various input types—sketches, scribbles, points, or boxes—all outperforming SAM and similar methods, validating the effectiveness of our network design. 



\subsection{Our Approach can be used for Annotation}
  Our method does more than just improve segmentation performance—it transforms a quantitative gain into a qualitative leap. The results generated by our approach can be directly used to train other models, just like GT obtained through pixel-by-pixel annotation. To demonstrate this, we applied our trained model to annotate the NC4K dataset \cite{lv2021simultaneously}, a newer camouflage segmentation dataset. By feeding the NC4K images and sketches drawn from the images into our network, we used the resulting labels (denoted as ``our predicted label" in Table \ref{table.pseudo}) as GT to train classic camouflage segmentation models like ZoomNet and UGTR. We then compared these models with those trained using real GT obtained from pixel-by-pixel annotation (denoted as ``pixel-by-pixel annotation"). To our surprise, the models trained with our labeled data performed on par with those trained with pixel-by-pixel annotations. An ANOVA (Analysis of Variance) confirmed the results were consistent (p $<$ 0.05). This demonstrates that our method can be effectively used for data annotation, providing a significant speed advantage of over 120 times faster than traditional pixel-wise annotation methods, which usually take around 60 minutes. (More details of the experiment in Supplementary Material).
  
\vskip -0.2in
\begin{table}[H]
\caption{Our method can be used for annotating new datasets and train other networks.}
\vskip 0.2in
\centering
\resizebox{1\linewidth}{!}{
\begin{tabular}{ll|llll}
\hline
\multicolumn{2}{c|}{Trained with label}                                                        & $S_m \uparrow$ & $E_{m} \uparrow$ & $F_\beta^w \uparrow$ & MAE$ \downarrow$ \\ \hline
\multicolumn{1}{c|}{\multirow{3}{*}{\parbox{2cm}{ZoomNet\newline \cite{zoomnet}}}}                           & Pixel-by-Pixel Annotated   & 0.818          & 0.884            & 0.688                & 0.034            \\ \cline{2-6} 
\multicolumn{1}{c|}{}   & Our Predicted Label & 0.805          & 0.878            & 0.673                & 0.036            \\  \cline{2-6} 
\multicolumn{1}{l|}{}                                                                  &      & \multicolumn{4}{c}{p-value = 0.0323, f-value = 4.5818}                       \\ \hline
\multicolumn{1}{c|}{\multirow{3}{*}{\parbox{2cm}{UGTR \cite{ugtr}}}}                                 & Pixel-by-Pixel Annotated   & 0.792          & 0.864            & 0.601                & 0.046            \\ \cline{2-6} 
\multicolumn{1}{c|}{}      & Our Predicted Label & 0.799          & 0.871            & 0.609                & 0.044            \\  \cline{2-6} 
\multicolumn{1}{l|}{}                                                                  &      & \multicolumn{4}{c}{p-value = 0.0068, f-value = 7.3182}                       \\ \hline
\end{tabular}}
\label{table.pseudo}
\vskip -0.1in
\end{table}

\begin{figure}[htbp]
    \centering
    \includegraphics[width=0.4\textwidth]{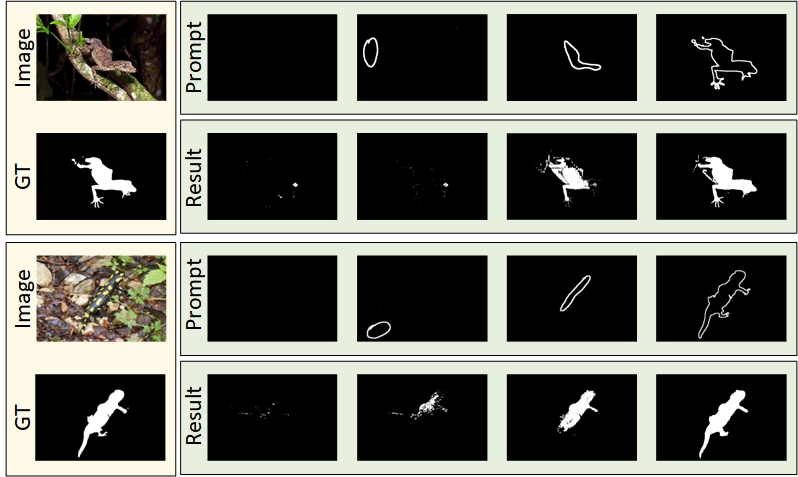}
    \vskip -0.1in
    \caption{Sketching the ``wrong" region. The experiment shows that our network has successfully used the sketch information as the key reference.}
    \vskip -0.2in
    \label{fig:ref_other}
  \end{figure}

\subsection{Ablation Study}
\noindent\textbf{Sketching Guidance.} To verify that the segmentation is guided by the sketch, we conduct an experiment by sketching the ``wrong region". The network fails to produce an accurate result when sketching the ``wrong label", showing that the sketch contributed to the segmentation process, not solely rely on the capability of SAM backbone, see in Figure \ref{fig:ref_other} (More visualization in Supplementary Material).

\textbf{Adapting Domain Specific Information.}
As shown in Table \ref{table.adapter}, injecting domain-specific information helps the performance boost. Details in Supplementary Material.

\vskip -0.1in
\begin{table}[H]
\vskip -0.2in
\caption{Ablation study on the domain-specific info (Inj. Info.).}

\centering
\normalsize
\resizebox{0.85\linewidth}{!}{
\begin{tabular}{ll|llll}
\hline
\multicolumn{2}{c|}{Method}           & \multicolumn{1}{c}{$S_m \uparrow$} & \multicolumn{1}{c}{$E_{m} \uparrow$} & \multicolumn{1}{c}{$F_\beta^w \uparrow$} & \multicolumn{1}{c}{MAE$ \downarrow$}                                                                                                                                      \\ \hline
\multicolumn{1}{l|}{Baseline} & Inj. Info.  & \multicolumn{4}{c}{COD10K}                                                                                                                                  \\ \hline
$\checkmark$                  & $\times$     & 0.867\scriptsize\normalsize                  & 0.950\scriptsize\normalsize                    & 0.807\scriptsize\normalsize                        & 0.023\scriptsize\normalsize                    \\
$\checkmark$                  & $\checkmark$ & 0.872\scriptsize\normalsize                  & 0.954\scriptsize\normalsize                    & 0.817\scriptsize\normalsize                        & 0.021\scriptsize\normalsize                        \\ \hline

\end{tabular}}

\label{table.adapter}
\vskip -0.1in
\end{table}

\noindent\textbf{Impact of Adaptive Focal Loss, Boundary Refinement and Sketch Augmentation.} Table \ref{table.ablation} indicates that Adaptive Focal Loss, Boundary Refinement, and Sketch Augmentation these design elements that contribute to performance improvement. Details in Supplementary Material.

\noindent\textbf{Impact of Sketch Augmentation and ``The better the drawings, the better the outcome".}
A key aspect of our sketch augmentation is the parameter $\Delta$ to simulate various potential freehand deviations, such as stretching, twisting, and rotation, so that we can obtain different sketches to enhance the robustness of the model. To verify the effectiveness of the sketch augmentation model, we compared the model trained w/ and w/o augmented data. It can be found that with the introduction of training data augmentation, the model's performance is enhanced. We further verified the different magnitude of perturbation of the test data (from the KOSCamo+ dataset). When applying strong perturbations to the test data with a trained model, performance declines, which matches our claim that the more accurate the drawing is, the better the performance (that makes intuitive sense). The results are shown in Table \ref{table.trainandsketch}.

\noindent\textbf{Sensitivity Analysis for Sketch Augmentation.}
Table \ref{table.sa_patch} demonstrates the hyperparameter patch quantity $n$'s effect on the performance. Details in Supplementary Material. 


\begin{table}[!htbp]
\vskip -0.2in
\caption{The effectiveness of Adaptive Focal Loss, Boundary Refinement and Sketch Augmentation.}
\resizebox{\linewidth}{!}{
\begin{tabular}{lll|llll}
\hline
\multicolumn{3}{c|}{Method}                                                              & \multicolumn{1}{c}{$S_m \uparrow$} & \multicolumn{1}{c}{$E_{m} \uparrow$} & \multicolumn{1}{c}{$F_\beta^w \uparrow$} & \multicolumn{1}{c}{MAE$ \downarrow$} 
\\ \hline
\multicolumn{1}{c|}{Adaptive} & \multicolumn{1}{c|}{Boundary} & \multicolumn{1}{c|}{SA} & \multicolumn{4}{c}{COD10K}                                                                                                                                  \\ \hline
$\times$                      & $\checkmark$                  & $\checkmark$             & 0.874\scriptsize                  & 0.955\scriptsize                 & 0.824\scriptsize                     & 0.020\scriptsize                 \\
$\checkmark$                  & $\times$                      & $\checkmark$             & 0.868\scriptsize                & 0.948\scriptsize                  & 0.813\scriptsize                      & 0.022\scriptsize                 \\
$\checkmark$                  & $\checkmark$                  & $\times$                 & 0.875\scriptsize                & 0.952\scriptsize                 & 0.825\scriptsize                      & 0.021\scriptsize                   \\
$\checkmark$                  & $\checkmark$                  & $\checkmark$             & 0.875\scriptsize               & 0.956\scriptsize                 & 0.825\scriptsize                      & 0.021\scriptsize \\
\hline

\end{tabular}}
\label{table.ablation}
\end{table}


\begin{table}[!htbp]
\vskip -0.3in
\caption{Sensitivity Analysis on $n$ in SA on COD10K.}
\centering  
\resizebox{0.6\linewidth}{!}{
\begin{tabular}{l|llll}
\hline
       $n$   & $S_{m}\uparrow$ & $E_{\text{m}} \uparrow$ & $F_{\beta}^{w} \uparrow$ & MAE$\downarrow$ \\ \hline
36 & 0.859  & 0.945          & 0.796           & 0.024  \\
64 & \textbf{0.875}  & \textbf{0.956}  & \textbf{0.825}  & \textbf{0.021}  \\ 
81 & 0.860  & 0.944  & 0.799  & 0.024  \\ \hline
\end{tabular}}
\label{table.sa_patch}
\vskip -0.1in
\end{table}


\begin{table}[!htbp]
\vskip -0.2in
\caption{Different Sketch Enhancement/Perturbations in training/testing data affect the performance.}
\centering
\resizebox{0.85\linewidth}{!}{
\begin{tabular}{cc|cccc}
\hline
Training data & Test Data & $S_m \uparrow$ & $E_{m} \uparrow$ & $F_\beta^w \uparrow$ & MAE$ \downarrow$ \\ \hline
Augmented     & -         & \textbf{0.876} & \textbf{0.957}   & \textbf{0.827}       & \textbf{0.021}   \\
Augmented     & Weak      & 0.872          & 0.952            & 0.818                & \textbf{0.021}   \\
Augmented     & Strong    & 0.871          & 0.950            & 0.815                & 0.022            \\
-             & -         & 0.875          & 0.952            & 0.825                & \textbf{0.021}   \\
-             & Weak      & 0.870          & 0.947            & 0.815                & 0.022            \\
-             & Strong    & 0.869          & 0.947            & 0.812                & 0.023            \\ \hline
\end{tabular}}
\label{table.trainandsketch}
\vskip -0.1in
\end{table}


\noindent\textbf{Impact of Adaptive Focal Loss $\theta$.} $\theta = (1-P_t^i)^{(\gamma+\gamma_\alpha)}$ is a difficulty modifier. The Figure \ref{fig:figure 3} presents the results for different values. Details in Supplementary Material.

\vskip -0.1in
\begin{figure}[H]
    \centering
    \begin{minipage}[b]{0.235\textwidth} 
    \centering
    \includegraphics[width=\linewidth]{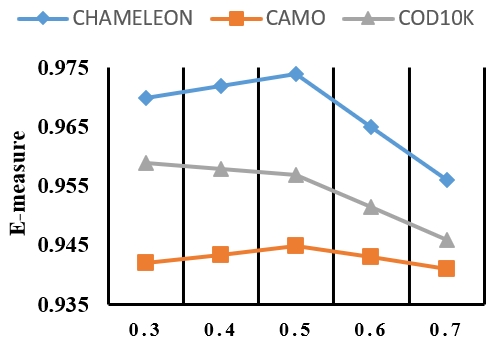}
    \vskip -0.1in
    \caption{The analysis of $E_m$ metrics on the COD10K dataset about $\theta$ in Adaptive Focal Loss.}
    \vskip -0.2in
    \label{fig:figure 3}
    \end{minipage}%
    \hfill
    \begin{minipage}[b]{0.235\textwidth} 
    \centering
    \includegraphics[width=1.12\linewidth]{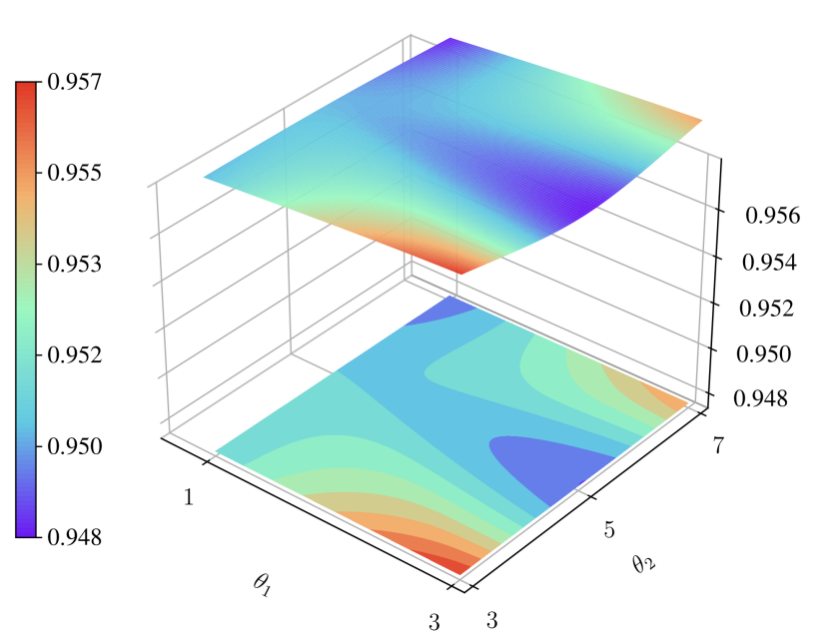}
    \vskip -0.1in
    \caption{Sensitivity Anallysis of $E_m$ on the COD10K dataset.}
    \vskip -0.2in
    \label{fig:merged iamge}
\end{minipage}
\end{figure}

\noindent\textbf{Impact of Boundary Refinement $\theta_1$ and $\theta_2$.} 
We investigate different combinations of $\theta_1$ and $\theta_2$ on performance (Figure \ref{fig:merged iamge}). Details in Supplementary Material.

\vspace{-0.2cm}
\section{Conclusion}
\vspace{-0.1cm}
Our work demonstrates the transformative potential of freehand sketches in challenging image segmentation tasks. We have developed a method that not only enhances segmentation accuracy but also significantly reduces annotation time. Our results show that sketches can outperform traditional prompting techniques for existing networks. Furthermore, we designed our network and gained an extra performance boost. Our network can produce segmentation results equivalent to ground truth masks while saving up to 120 times the annotation effort underscores its practical utility. We believe that this research paves the way for more efficient and user-friendly computer vision applications, bridging the gap between human creativity and machine intelligence. 

\section*{Impact Statement}
This paper proposes an innovative sketch-guided interactive image segmentation framework, aimed at enhancing segmentation performance through human freehand sketches and facilitate data annotation. We believe that this research does not involve any ethical issues and will not have any negative impact on user privacy or data security.

\nocite{langley00}

\bibliography{example_paper}

\begin{thebibliography}{85}
\providecommand{\natexlab}[1]{#1}
\providecommand{\url}[1]{\texttt{#1}}
\expandafter\ifx\csname urlstyle\endcsname\relax
  \providecommand{\doi}[1]{doi: #1}\else
  \providecommand{\doi}{doi: \begingroup \urlstyle{rm}\Url}\fi

\bibitem[Bhunia et~al.(2020)Bhunia, Das, Muhammad, Yang, Hospedales, Xiang, Gryaditskaya, and Song]{bhunia2020pixelor}
Bhunia, A.~K., Das, A., Muhammad, U.~R., Yang, Y., Hospedales, T.~M., Xiang, T., Gryaditskaya, Y., and Song, Y.-Z.
\newblock Pixelor: A competitive sketching ai agent. so you think you can sketch?
\newblock \emph{ACM Transactions on Graphics (TOG)}, 39\penalty0 (6):\penalty0 1--15, 2020.

\bibitem[Bhunia et~al.(2023)Bhunia, Koley, Kumar, Sain, Chowdhury, Xiang, and Song]{bhunia2023sketch2saliency}
Bhunia, A.~K., Koley, S., Kumar, A., Sain, A., Chowdhury, P.~N., Xiang, T., and Song, Y.-Z.
\newblock Sketch2saliency: learning to detect salient objects from human drawings.
\newblock In \emph{Proceedings of the IEEE/CVF conference on computer vision and pattern recognition}, pp.\  2733--2743, 2023.

\bibitem[Bokhovkin \& Burnaev(2019)Bokhovkin and Burnaev]{bokhovkin2019boundary}
Bokhovkin, A. and Burnaev, E.
\newblock Boundary loss for remote sensing imagery semantic segmentation.
\newblock In \emph{International Symposium on Neural Networks}, pp.\  388--401. Springer, 2019.

\bibitem[Chen et~al.(2025)Chen, Shao, Guo, and Gao]{chen2025hint}
Chen, H., Shao, D., Guo, G., and Gao, S.
\newblock Just a hint: Point-supervised camouflaged object detection.
\newblock In \emph{European Conference on Computer Vision}, pp.\  332--348. Springer, Cham, 2025.

\bibitem[Chen et~al.(2017)Chen, Papandreou, Kokkinos, Murphy, and Yuille]{chen2017deeplab}
Chen, L.-C., Papandreou, G., Kokkinos, I., Murphy, K., and Yuille, A.~L.
\newblock Deeplab: Semantic image segmentation with deep convolutional nets, atrous convolution, and fully connected crfs.
\newblock \emph{IEEE transactions on pattern analysis and machine intelligence}, 40\penalty0 (4):\penalty0 834--848, 2017.

\bibitem[Chen et~al.(2023{\natexlab{a}})Chen, Ding, Zhu, Zang, Liao, Li, and Sun]{chen2023reality3dsketch}
Chen, T., Ding, C., Zhu, L., Zang, Y., Liao, Y., Li, Z., and Sun, L.
\newblock Reality3dsketch: rapid 3d modeling of objects from single freehand sketches.
\newblock \emph{IEEE Transactions on Multimedia}, 2023{\natexlab{a}}.

\bibitem[Chen et~al.(2023{\natexlab{b}})Chen, Fu, Zhu, Mao, Zhang, Zang, and Sun]{deep3dsketch}
Chen, T., Fu, C., Zhu, L., Mao, P., Zhang, J., Zang, Y., and Sun, L.
\newblock Deep3dsketch: 3d modeling from free-hand sketches with view-and structural-aware adversarial training.
\newblock \emph{arXiv preprint arXiv:2312.04435}, 2023{\natexlab{b}}.

\bibitem[Chen et~al.(2023{\natexlab{c}})Chen, Zhu, Deng, Cao, Wang, Zhang, Li, Sun, Zang, and Mao]{sam-adapter}
Chen, T., Zhu, L., Deng, C., Cao, R., Wang, Y., Zhang, S., Li, Z., Sun, L., Zang, Y., and Mao, P.
\newblock Sam-adapter: Adapting segment anything in underperformed scenes.
\newblock In \emph{Proceedings of the IEEE/CVF International Conference on Computer Vision}, pp.\  3367--3375, 2023{\natexlab{c}}.

\bibitem[Chen et~al.(2024{\natexlab{a}})Chen, Cao, Li, Zang, and Sun]{deep3dsketch-im}
Chen, T., Cao, R., Li, Z., Zang, Y., and Sun, L.
\newblock Deep3dsketch-im: rapid high-fidelity ai 3d model generation by single freehand sketches.
\newblock \emph{Frontiers of Information Technology \& Electronic Engineering}, 25\penalty0 (1):\penalty0 149--159, 2024{\natexlab{a}}.

\bibitem[Chen et~al.(2024{\natexlab{b}})Chen, Cao, Lu, Xu, Zhang, Papa, Zhang, Sun, and Zang]{chen2024high}
Chen, T., Cao, R., Lu, A., Xu, T., Zhang, X., Papa, M., Zhang, M., Sun, L., and Zang, Y.
\newblock High-fidelity 3d model generation with relightable appearance from single freehand sketches and text guidance.
\newblock In \emph{2024 IEEE International Conference on Multimedia and Expo Workshops (ICMEW)}, pp.\  1--6. IEEE, 2024{\natexlab{b}}.

\bibitem[Chen et~al.(2024{\natexlab{c}})Chen, Chen, Ding, Bai, Zhang, Zhu, Zang, Hu, Li, and Sun]{chen2024new}
Chen, T., Chen, X., Ding, C., Bai, L., Zhang, S., Zhu, L., Zang, Y., Hu, W., Li, Z., and Sun, L.
\newblock New fashion: Personalized 3d design with a single sketch input.
\newblock In \emph{SIGGRAPH Asia 2024 Posters}, pp.\  1--3. 2024{\natexlab{c}}.

\bibitem[Chen et~al.(2024{\natexlab{d}})Chen, Ding, Zhang, Yu, Zang, Li, Peng, and Sun]{Deep3DVRSketch}
Chen, T., Ding, C., Zhang, S., Yu, C., Zang, Y., Li, Z., Peng, S., and Sun, L.
\newblock Rapid 3d model generation with intuitive 3d input.
\newblock In \emph{Proceedings of the IEEE/CVF Conference on Computer Vision and Pattern Recognition}, pp.\  12554--12564, 2024{\natexlab{d}}.

\bibitem[Chen et~al.(2024{\natexlab{e}})Chen, Ding, Zhang, Yu, Zang, Li, Peng, and Sun]{chen2024rapid}
Chen, T., Ding, C., Zhang, S., Yu, C., Zang, Y., Li, Z., Peng, S., and Sun, L.
\newblock Rapid 3d model generation with intuitive 3d input.
\newblock In \emph{Proceedings of the IEEE/CVF Conference on Computer Vision and Pattern Recognition}, pp.\  12554--12564, 2024{\natexlab{e}}.

\bibitem[Chen et~al.(2024{\natexlab{f}})Chen, Lu, Zhu, Ding, Yu, Ji, Li, Sun, Mao, and Zang]{chen2024sam2}
Chen, T., Lu, A., Zhu, L., Ding, C., Yu, C., Ji, D., Li, Z., Sun, L., Mao, P., and Zang, Y.
\newblock Sam2-adapter: Evaluating \& adapting segment anything 2 in downstream tasks: Camouflage, shadow, medical image segmentation, and more.
\newblock \emph{arXiv preprint arXiv:2408.04579}, 2024{\natexlab{f}}.

\bibitem[Chen et~al.(2024{\natexlab{g}})Chen, Yu, Hu, Li, Xu, Cao, Zhu, Zang, Zhang, Li, et~al.]{chen2024img2cad}
Chen, T., Yu, C., Hu, Y., Li, J., Xu, T., Cao, R., Zhu, L., Zang, Y., Zhang, Y., Li, Z., et~al.
\newblock Img2cad: Conditioned 3d cad model generation from single image with structured visual geometry.
\newblock \emph{arXiv preprint arXiv:2410.03417}, 2024{\natexlab{g}}.

\bibitem[Chen et~al.(2024{\natexlab{h}})Chen, Yu, Li, Zhang, Zhu, Ji, Zhang, Zang, Li, and Sun]{chen2024reasoning3d}
Chen, T., Yu, C., Li, J., Zhang, J., Zhu, L., Ji, D., Zhang, Y., Zang, Y., Li, Z., and Sun, L.
\newblock Reasoning3d--grounding and reasoning in 3d: Fine-grained zero-shot open-vocabulary 3d reasoning part segmentation via large vision-language models.
\newblock \emph{arXiv preprint arXiv:2405.19326}, 2024{\natexlab{h}}.

\bibitem[Chowdhury et~al.(2023)Chowdhury, Bhunia, Sain, Koley, Xiang, and Song]{whathumancan}
Chowdhury, P.~N., Bhunia, A.~K., Sain, A., Koley, S., Xiang, T., and Song, Y.-Z.
\newblock What can human sketches do for object detection?
\newblock In \emph{Proceedings of the IEEE/CVF conference on computer vision and pattern recognition}, pp.\  15083--15094, 2023.

\bibitem[Del~Bimbo \& Pala(1997)Del~Bimbo and Pala]{visualimage}
Del~Bimbo, A. and Pala, P.
\newblock Visual image retrieval by elastic matching of user sketches.
\newblock \emph{IEEE Transactions on Pattern Analysis and Machine Intelligence}, 19\penalty0 (2):\penalty0 121--132, 1997.

\bibitem[Ding et~al.(2020)Ding, Cohen, Price, and Jiang]{new7}
Ding, H., Cohen, S., Price, B., and Jiang, X.
\newblock Phraseclick: toward achieving flexible interactive segmentation by phrase and click.
\newblock In \emph{Computer Vision--ECCV 2020: 16th European Conference, Glasgow, UK, August 23--28, 2020, Proceedings, Part III 16}, pp.\  417--435. Springer, 2020.

\bibitem[Ding et~al.(2021)Ding, Liu, Wang, and Jiang]{3-2}
Ding, H., Liu, C., Wang, S., and Jiang, X.
\newblock Vision-language transformer and query generation for referring segmentation.
\newblock In \emph{Proceedings of the IEEE/CVF International Conference on Computer Vision (ICCV)}, pp.\  16321--16330, 2021.

\bibitem[Eitz et~al.(2010)Eitz, Hildebrand, Boubekeur, and Alexa]{sketch-based}
Eitz, M., Hildebrand, K., Boubekeur, T., and Alexa, M.
\newblock Sketch-based image retrieval: Benchmark and bag-of-features descriptors.
\newblock \emph{IEEE transactions on visualization and computer graphics}, 17\penalty0 (11):\penalty0 1624--1636, 2010.

\bibitem[Esmaeilzehi et~al.(2021)Esmaeilzehi, Ahmad, and Swamy]{fpnet}
Esmaeilzehi, A., Ahmad, M.~O., and Swamy, M.
\newblock Fpnet: A deep light-weight interpretable neural network using forward prediction filtering for efficient single image super resolution.
\newblock \emph{IEEE Transactions on Circuits and Systems II: Express Briefs}, 69\penalty0 (3):\penalty0 1937--1941, 2021.

\bibitem[Fan et~al.(2020)Fan, Ji, Sun, Cheng, Shen, and Shao]{sinet}
Fan, D.-P., Ji, G.-P., Sun, G., Cheng, M.-M., Shen, J., and Shao, L.
\newblock Camouflaged object detection.
\newblock In \emph{Proceedings of the IEEE/CVF conference on computer vision and pattern recognition}, pp.\  2777--2787, 2020.

\bibitem[Fioraldi et~al.(2023)Fioraldi, Mantovani, Maier, and Balzarotti]{aml}
Fioraldi, A., Mantovani, A., Maier, D., and Balzarotti, D.
\newblock Dissecting american fuzzy lop: a fuzzbench evaluation.
\newblock \emph{ACM transactions on software engineering and methodology}, 32\penalty0 (2):\penalty0 1--26, 2023.

\bibitem[Goodwin et~al.(2007)Goodwin, Vollick, and Hertzmann]{goodwin2007isophote}
Goodwin, T., Vollick, I., and Hertzmann, A.
\newblock Isophote distance: a shading approach to artistic stroke thickness.
\newblock In \emph{Proceedings of the 5th international symposium on Non-photorealistic animation and rendering}, pp.\  53--62, 2007.

\bibitem[Grady(2006)]{grady2006random}
Grady, L.
\newblock Random walks for image segmentation.
\newblock \emph{IEEE transactions on pattern analysis and machine intelligence}, 28\penalty0 (11):\penalty0 1768--1783, 2006.

\bibitem[He et~al.(2023{\natexlab{a}})He, Li, Zhang, Tang, Zhang, Guo, and Li]{feder}
He, C., Li, K., Zhang, Y., Tang, L., Zhang, Y., Guo, Z., and Li, X.
\newblock Camouflaged object detection with feature decomposition and edge reconstruction.
\newblock In \emph{Proceedings of the IEEE/CVF conference on computer vision and pattern recognition}, pp.\  22046--22055, 2023{\natexlab{a}}.

\bibitem[He et~al.(2024)He, Li, Zhang, Xu, Tang, Zhang, and Li]{he2024weakly}
He, C., Li, K., Zhang, Y., Xu, G., Tang, L., Zhang, Y., and Li, X.
\newblock Weakly-supervised concealed object segmentation with sam-based pseudo labeling and multi-scale feature grouping.
\newblock \emph{Advances in Neural Information Processing Systems}, 36, 2024.

\bibitem[He et~al.(2016)He, Zhang, Ren, and Sun]{resnet}
He, K., Zhang, X., Ren, S., and Sun, J.
\newblock Deep residual learning for image recognition.
\newblock In \emph{Proceedings of the IEEE conference on computer vision and pattern recognition}, pp.\  770--778, 2016.

\bibitem[He et~al.(2023{\natexlab{b}})He, Dong, Lin, and Lau]{he2023weakly}
He, R., Dong, Q., Lin, J., and Lau, R.~W.
\newblock Weakly-supervised camouflaged object detection with scribble annotations.
\newblock In \emph{Proceedings of the AAAI Conference on Artificial Intelligence}, volume~37, pp.\  781--789, 2023{\natexlab{b}}.

\bibitem[Hendrycks \& Gimpel(2016)Hendrycks and Gimpel]{hendrycks2016gaussian}
Hendrycks, D. and Gimpel, K.
\newblock Gaussian error linear units (gelus).
\newblock \emph{arXiv preprint arXiv:1606.08415}, 2016.

\bibitem[Hertzmann(2020)]{1}
Hertzmann, A.
\newblock Why do line drawings work? a realism hypothesis.
\newblock \emph{Perception}, 49\penalty0 (4):\penalty0 439--451, 2020.

\bibitem[Hu et~al.(2020{\natexlab{a}})Hu, Li, Yang, Hospedales, and Song]{hu2020sketch}
Hu, C., Li, D., Yang, Y., Hospedales, T.~M., and Song, Y.-Z.
\newblock Sketch-a-segmenter: Sketch-based photo segmenter generation.
\newblock \emph{IEEE transactions on image processing}, 29:\penalty0 9470--9481, 2020{\natexlab{a}}.

\bibitem[Hu et~al.(2016)Hu, Rohrbach, and Darrell]{3-1}
Hu, R., Rohrbach, M., and Darrell, T.
\newblock Segmentation from natural language expressions.
\newblock In \emph{Computer Vision–ECCV 2016: 14th European Conference, Amsterdam, The Netherlands, October 11–14, 2016, Proceedings, Part I 14}, pp.\  108--124. Springer, 2016.

\bibitem[Hu et~al.(2024)Hu, Zhang, Bai, Li, Li, Zang, and Hu]{hu2024sketch}
Hu, Y., Zhang, J., Bai, L., Li, J., Li, B., Zang, Y., and Hu, W.
\newblock From sketch to reality: precision-friendly 3d generation technology.
\newblock \emph{The Visual Computer}, pp.\  1--12, 2024.

\bibitem[Hu et~al.(2020{\natexlab{b}})Hu, Feng, Sun, Zhang, and Lu]{uniref++35}
Hu, Z., Feng, G., Sun, J., Zhang, L., and Lu, H.
\newblock Bi-directional relationship inferring network for referring image segmentation.
\newblock In \emph{Proceedings of the IEEE/CVF conference on computer vision and pattern recognition}, pp.\  4424--4433, 2020{\natexlab{b}}.

\bibitem[Huang et~al.(2023)Huang, Dai, Xiang, Wang, Chen, Qin, and Xiong]{fspnet}
Huang, Z., Dai, H., Xiang, T.-Z., Wang, S., Chen, H.-X., Qin, J., and Xiong, H.
\newblock Feature shrinkage pyramid for camouflaged object detection with transformers.
\newblock In \emph{Proceedings of the IEEE/CVF conference on computer vision and pattern recognition}, pp.\  5557--5566, 2023.

\bibitem[Kennedy(1974)]{kennedy1974psychology}
Kennedy, J.
\newblock A psychology of picture perception, 1974.

\bibitem[Kirillov et~al.(2023{\natexlab{a}})Kirillov, Mintun, Ravi, Mao, Rolland, Gustafson, Xiao, Whitehead, Berg, Lo, et~al.]{kirillov2023segment}
Kirillov, A., Mintun, E., Ravi, N., Mao, H., Rolland, C., Gustafson, L., Xiao, T., Whitehead, S., Berg, A.~C., Lo, W.-Y., et~al.
\newblock Segment anything.
\newblock In \emph{Proceedings of the IEEE/CVF International Conference on Computer Vision}, pp.\  4015--4026, 2023{\natexlab{a}}.

\bibitem[Kirillov et~al.(2023{\natexlab{b}})Kirillov, Mintun, Ravi, Mao, Rolland, Gustafson, Xiao, Whitehead, Berg, Lo, et~al.]{sam}
Kirillov, A., Mintun, E., Ravi, N., Mao, H., Rolland, C., Gustafson, L., Xiao, T., Whitehead, S., Berg, A.~C., Lo, W.-Y., et~al.
\newblock Segment anything.
\newblock In \emph{Proceedings of the IEEE/CVF International Conference on Computer Vision}, pp.\  4015--4026, 2023{\natexlab{b}}.

\bibitem[Koley et~al.(2024{\natexlab{a}})Koley, Bhunia, Sain, Chowdhury, Xiang, and Song]{8}
Koley, S., Bhunia, A.~K., Sain, A., Chowdhury, P.~N., Xiang, T., and Song, Y.-Z.
\newblock How to handle sketch-abstraction in sketch-based image retrieval?
\newblock In \emph{Proceedings of the IEEE/CVF Conference on Computer Vision and Pattern Recognition}, pp.\  16859--16869, 2024{\natexlab{a}}.

\bibitem[Koley et~al.(2024{\natexlab{b}})Koley, Bhunia, Sekhri, Sain, Chowdhury, Xiang, and Song]{koley2024s}
Koley, S., Bhunia, A.~K., Sekhri, D., Sain, A., Chowdhury, P.~N., Xiang, T., and Song, Y.-Z.
\newblock It's all about your sketch: Democratising sketch control in diffusion models.
\newblock In \emph{Proceedings of the IEEE/CVF Conference on Computer Vision and Pattern Recognition}, pp.\  7204--7214, 2024{\natexlab{b}}.

\bibitem[Le et~al.(2019)Le, Nguyen, Nie, Tran, and Sugimoto]{sam-adapter27}
Le, T.-N., Nguyen, T.~V., Nie, Z., Tran, M.-T., and Sugimoto, A.
\newblock Anabranch network for camouflaged object segmentation.
\newblock \emph{Computer Vision and Image Understanding}, 184:\penalty0 45--56, 2019.

\bibitem[Li et~al.(2021)Li, Zhang, Lv, Liu, Zhang, and Dai]{ujsc}
Li, A., Zhang, J., Lv, Y., Liu, B., Zhang, T., and Dai, Y.
\newblock Uncertainty-aware joint salient object and camouflaged object detection.
\newblock In \emph{Proceedings of the IEEE/CVF conference on computer vision and pattern recognition}, pp.\  10071--10081, 2021.

\bibitem[Liu et~al.(2022{\natexlab{a}})Liu, Xu, Jiao, and Niethammer]{new25}
Liu, Q., Xu, Z., Jiao, Y., and Niethammer, M.
\newblock isegformer: interactive segmentation via transformers with application to 3d knee mr images.
\newblock In \emph{International Conference on Medical Image Computing and Computer-Assisted Intervention}, pp.\  464--474. Springer, 2022{\natexlab{a}}.

\bibitem[Liu et~al.(2022{\natexlab{b}})Liu, Zheng, Planche, Karanam, Chen, Niethammer, and Wu]{new26}
Liu, Q., Zheng, M., Planche, B., Karanam, S., Chen, T., Niethammer, M., and Wu, Z.
\newblock Pseudoclick: Interactive image segmentation with click imitation.
\newblock In \emph{European Conference on Computer Vision}, pp.\  728--745. Springer, 2022{\natexlab{b}}.

\bibitem[Liu et~al.(2023{\natexlab{a}})Liu, Xu, Bertasius, and Niethammer]{new27}
Liu, Q., Xu, Z., Bertasius, G., and Niethammer, M.
\newblock Simpleclick: Interactive image segmentation with simple vision transformers.
\newblock In \emph{Proceedings of the IEEE/CVF International Conference on Computer Vision}, pp.\  22290--22300, 2023{\natexlab{a}}.

\bibitem[Liu et~al.(2023{\natexlab{b}})Liu, Shen, Pun, and Cun]{evp}
Liu, W., Shen, X., Pun, C.-M., and Cun, X.
\newblock Explicit visual prompting for low-level structure segmentations.
\newblock In \emph{Proceedings of the IEEE/CVF Conference on Computer Vision and Pattern Recognition}, pp.\  19434--19445, 2023{\natexlab{b}}.

\bibitem[Lv et~al.(2021{\natexlab{a}})Lv, Zhang, Dai, Li, Liu, Barnes, and Fan]{4-29}
Lv, Y., Zhang, J., Dai, Y., Li, A., Liu, B., Barnes, N., and Fan, D.-P.
\newblock Simultaneously localize, segment and rank the camouflaged objects.
\newblock In \emph{Computer Vision and Pattern Recognition}, pp.\  11591--115601, 2021{\natexlab{a}}.

\bibitem[Lv et~al.(2021{\natexlab{b}})Lv, Zhang, Dai, Li, Liu, Barnes, and Fan]{lv2021simultaneously}
Lv, Y., Zhang, J., Dai, Y., Li, A., Liu, B., Barnes, N., and Fan, D.-P.
\newblock Simultaneously localize, segment and rank the camouflaged objects.
\newblock In \emph{Proceedings of the IEEE/CVF conference on computer vision and pattern recognition}, pp.\  11591--11601, 2021{\natexlab{b}}.

\bibitem[Matsui et~al.(2017)Matsui, Ito, Aramaki, Fujimoto, Ogawa, Yamasaki, and Aizawa]{sketch-basedmanged}
Matsui, Y., Ito, K., Aramaki, Y., Fujimoto, A., Ogawa, T., Yamasaki, T., and Aizawa, K.
\newblock Sketch-based manga retrieval using manga109 dataset.
\newblock \emph{Multimedia tools and applications}, 76:\penalty0 21811--21838, 2017.

\bibitem[Mei et~al.(2021)Mei, Ji, Wei, Yang, Wei, and Fan]{pfnet}
Mei, H., Ji, G.-P., Wei, Z., Yang, X., Wei, X., and Fan, D.-P.
\newblock Camouflaged object segmentation with distraction mining.
\newblock In \emph{Proceedings of the IEEE/CVF Conference on Computer Vision and Pattern Recognition}, pp.\  8772--8781, 2021.

\bibitem[Olsen et~al.(2009)Olsen, Samavati, Sousa, and Jorge]{olsen2009sketch}
Olsen, L., Samavati, F.~F., Sousa, M.~C., and Jorge, J.~A.
\newblock Sketch-based modeling: A survey.
\newblock \emph{Computers \& Graphics}, 33\penalty0 (1):\penalty0 85--103, 2009.

\bibitem[Pan et~al.(2023)Pan, Yu, He, and Shi]{humansketch}
Pan, L., Yu, C., He, Z., and Shi, Y.
\newblock A human-computer collaborative editing tool for conceptual diagrams.
\newblock In \emph{Proceedings of the 2023 CHI Conference on Human Factors in Computing Systems}, pp.\  1--29, 2023.

\bibitem[Pang et~al.(2022{\natexlab{a}})Pang, Zhao, Xiang, Zhang, and Lu]{4-18}
Pang, Y., Zhao, X., Xiang, T., Zhang, L., and Lu, H.
\newblock Zoom in and out: A mixed-scale triplet network for camouflaged object detection.
\newblock In \emph{IEEE Conference on Computer Vision and Pattern Recognition}, pp.\  2150--2160, 2022{\natexlab{a}}.

\bibitem[Pang et~al.(2022{\natexlab{b}})Pang, Zhao, Xiang, Zhang, and Lu]{zoomnet}
Pang, Y., Zhao, X., Xiang, T.-Z., Zhang, L., and Lu, H.
\newblock Zoom in and out: A mixed-scale triplet network for camouflaged object detection.
\newblock In \emph{Proceedings of the IEEE/CVF Conference on computer vision and pattern recognition}, pp.\  2160--2170, 2022{\natexlab{b}}.

\bibitem[Pang et~al.(2023)Pang, Zhao, Zuo, Zhang, and Lu]{pang2023open}
Pang, Y., Zhao, X., Zuo, J., Zhang, L., and Lu, H.
\newblock Open-vocabulary camouflaged object segmentation.
\newblock \emph{arXiv preprint arXiv:2311.11241}, 2023.

\bibitem[Ravi et~al.(2024)Ravi, Gabeur, Hu, Hu, Ryali, Ma, Khedr, R{\"a}dle, Rolland, Gustafson, et~al.]{ravi2024sam}
Ravi, N., Gabeur, V., Hu, Y.-T., Hu, R., Ryali, C., Ma, T., Khedr, H., R{\"a}dle, R., Rolland, C., Gustafson, L., et~al.
\newblock Sam 2: Segment anything in images and videos.
\newblock \emph{arXiv preprint arXiv:2408.00714}, 2024.

\bibitem[Riba et~al.(2021)Riba, Dey, Biten, and Llados]{19}
Riba, P., Dey, S., Biten, A.~F., and Llados, J.
\newblock Localizing infinity-shaped fishes: Sketch-guided object localization in the wild.
\newblock \emph{arXiv preprint arXiv:2109.11874}, 2021.

\bibitem[Rozemberczki et~al.(2021)Rozemberczki, Allen, and Sarkar]{chamelon}
Rozemberczki, B., Allen, C., and Sarkar, R.
\newblock Multi-scale attributed node embedding, 2021.

\bibitem[Sangkloy et~al.(2022)Sangkloy, Jitkrittum, Yang, and Hays]{sangkloy2022sketch}
Sangkloy, P., Jitkrittum, W., Yang, D., and Hays, J.
\newblock A sketch is worth a thousand words: Image retrieval with text and sketch.
\newblock In \emph{European Conference on Computer Vision}, pp.\  251--267. Springer, 2022.

\bibitem[Sayim \& Cavanagh(2011)Sayim and Cavanagh]{sayim2011line}
Sayim, B. and Cavanagh, P.
\newblock What line drawings reveal about the visual brain.
\newblock \emph{Frontiers in human neuroscience}, 5:\penalty0 118, 2011.

\bibitem[Sengottuvelan et~al.(2008)Sengottuvelan, Wahi, and Shanmugam]{sam-adapter42}
Sengottuvelan, P., Wahi, A., and Shanmugam, A.
\newblock Performance of decamouflaging through exploratory image analysis.
\newblock In \emph{2008 First International Conference on Emerging Trends in Engineering and Technology}, pp.\  6--10. IEEE, 2008.

\bibitem[Shaheen et~al.(2017)Shaheen, Affara, and Ghanem]{shaheen2017constrained}
Shaheen, S., Affara, L., and Ghanem, B.
\newblock Constrained convolutional sparse coding for parametric based reconstruction of line drawings.
\newblock In \emph{Proceedings of the IEEE International Conference on Computer Vision}, pp.\  4424--4432, 2017.

\bibitem[Song et~al.(2017)Song, Song, Xiang, and Hospedales]{song2017fine}
Song, J., Song, Y.-Z., Xiang, T., and Hospedales, T.~M.
\newblock Fine-grained image retrieval: The text/sketch input dilemma.
\newblock In \emph{The 28th British Machine Vision Conference}, 2017.

\bibitem[Suzek et~al.(2007)Suzek, Huang, McGarvey, Mazumder, and Wu]{uniref}
Suzek, B.~E., Huang, H., McGarvey, P., Mazumder, R., and Wu, C.~H.
\newblock Uniref: comprehensive and non-redundant uniprot reference clusters.
\newblock \emph{Bioinformatics}, 23\penalty0 (10):\penalty0 1282--1288, 2007.

\bibitem[Tripathi et~al.(2020)Tripathi, Dani, Mishra, and Chakraborty]{20}
Tripathi, A., Dani, R.~R., Mishra, A., and Chakraborty, A.
\newblock Sketch-guided object localization in natural images.
\newblock In \emph{ECCV}, 2020.

\bibitem[Wang et~al.(2022)Wang, Lu, Li, Tao, Guo, Gong, and Liu]{uniref++99}
Wang, Z., Lu, Y., Li, Q., Tao, X., Guo, Y., Gong, M., and Liu, T.
\newblock Cris: Clip-driven referring image segmentation.
\newblock In \emph{Proceedings of the IEEE/CVF conference on computer vision and pattern recognition}, pp.\  11686--11695, 2022.

\bibitem[Wang et~al.(2023)Wang, Zhao, Xing, Xu, Kong, and Zhou]{3-4}
Wang, Z., Zhao, Z., Xing, X., Xu, D., Kong, X., and Zhou, L.
\newblock Conflict-based cross-view consistency for semi-supervised semantic segmentation.
\newblock In \emph{Proceedings of the IEEE/CVF Conference on Computer Vision and Pattern Recognition (CVPR)}, pp.\  19585--19595, 2023.

\bibitem[Wu et~al.(2023)Wu, Jiang, Yan, Lu, Yuan, and Luo]{uniref++}
Wu, J., Jiang, Y., Yan, B., Lu, H., Yuan, Z., and Luo, P.
\newblock Uniref++: Segment every reference object in spatial and temporal spaces.
\newblock \emph{arXiv preprint arXiv:2312.15715}, 2023.

\bibitem[Yang et~al.(2021)Yang, Zhai, Li, Huang, Luo, Cheng, and Fan]{ugtr}
Yang, F., Zhai, Q., Li, X., Huang, R., Luo, A., Cheng, H., and Fan, D.-P.
\newblock Uncertainty-guided transformer reasoning for camouflaged object detection.
\newblock In \emph{Proceedings of the IEEE/CVF international conference on computer vision}, pp.\  4146--4155, 2021.

\bibitem[Yang et~al.(2022)Yang, Wang, Tang, Chen, Zhao, and Torr]{3-3}
Yang, Z., Wang, J., Tang, Y., Chen, K., Zhao, H., and Torr, P.~H.
\newblock Lavt: Language-aware vision transformer for referring image segmentation.
\newblock In \emph{Proceedings of the IEEE/CVF Conference on Computer Vision and Pattern Recognition (CVPR)}, pp.\  18155--18165, 2022.

\bibitem[Yu et~al.(2018)Yu, Lin, Shen, Yang, Lu, Bansal, and Berg]{uniref++121}
Yu, L., Lin, Z., Shen, X., Yang, J., Lu, X., Bansal, M., and Berg, T.~L.
\newblock Mattnet: Modular attention network for referring expression comprehension.
\newblock In \emph{Proceedings of the IEEE conference on computer vision and pattern recognition}, pp.\  1307--1315, 2018.

\bibitem[Zang et~al.(2023{\natexlab{a}})Zang, Fu, Chen, Hu, Liu, and Hu]{deep3dsketch+}
Zang, Y., Fu, C., Chen, T., Hu, Y., Liu, Q., and Hu, W.
\newblock Deep3dsketch+: obtaining customized 3d model by single free-hand sketch through deep learning.
\newblock \emph{arXiv preprint arXiv:2310.18609}, 2023{\natexlab{a}}.

\bibitem[Zang et~al.(2023{\natexlab{b}})Zang, Fu, Chen, Hu, Liu, and Hu]{zang2023deep3dsketch+}
Zang, Y., Fu, C., Chen, T., Hu, Y., Liu, Q., and Hu, W.
\newblock Deep3dsketch+: obtaining customized 3d model by single free-hand sketch through deep learning.
\newblock \emph{arXiv preprint arXiv:2310.18609}, 2023{\natexlab{b}}.

\bibitem[Zang et~al.(2024)Zang, Han, Ding, Zhang, and Chen]{magic3dsketch}
Zang, Y., Han, Y., Ding, C., Zhang, J., and Chen, T.
\newblock Magic3dsketch: Create colorful 3d models from sketch-based 3d modeling guided by text and language-image pre-training.
\newblock \emph{arXiv preprint arXiv:2407.19225}, 2024.

\bibitem[Zang et~al.(2025)Zang, Cao, Fu, Zhu, Zhang, Hu, Zhu, and Chen]{zang2025resmatch}
Zang, Y., Cao, R., Fu, C., Zhu, D., Zhang, M., Hu, W., Zhu, L., and Chen, T.
\newblock Resmatch: Referring expression segmentation in a semi-supervised manner.
\newblock \emph{Information Sciences}, 694:\penalty0 121709, 2025.

\bibitem[Zhang et~al.(2020{\natexlab{a}})Zhang, Liew, Wei, Wei, and Zhao]{new45}
Zhang, S., Liew, J.~H., Wei, Y., Wei, S., and Zhao, Y.
\newblock Interactive object segmentation with inside-outside guidance.
\newblock In \emph{Proceedings of the IEEE/CVF conference on computer vision and pattern recognition}, pp.\  12234--12244, 2020{\natexlab{a}}.

\bibitem[Zhang et~al.(2020{\natexlab{b}})Zhang, Liew, Wei, Wei, and Zhao]{zhang2020interactive}
Zhang, S., Liew, J.~H., Wei, Y., Wei, S., and Zhao, Y.
\newblock Interactive object segmentation with inside-outside guidance.
\newblock In \emph{Proceedings of the IEEE/CVF conference on computer vision and pattern recognition}, pp.\  12234--12244, 2020{\natexlab{b}}.

\bibitem[Zhang et~al.(2023)Zhang, Yin, Lin, Hou, Fan, and Cheng]{refcod}
Zhang, X., Yin, B., Lin, Z., Hou, Q., Fan, D.-P., and Cheng, M.-M.
\newblock Referring camouflaged object detection.
\newblock \emph{arXiv preprint arXiv:2306.07532}, 2023.

\bibitem[Zheng et~al.(2021)Zheng, Yao, Sun, Zhang, Zhao, and Porikli]{zheng2021sketch}
Zheng, Y., Yao, H., Sun, X., Zhang, S., Zhao, S., and Porikli, F.
\newblock Sketch-specific data augmentation for freehand sketch recognition.
\newblock \emph{Neurocomputing}, 456:\penalty0 528--539, 2021.

\bibitem[Zhu et~al.(2021)Zhu, Ji, Zhu, Gan, Wu, and Yan]{zhu2021learning}
Zhu, L., Ji, D., Zhu, S., Gan, W., Wu, W., and Yan, J.
\newblock Learning statistical texture for semantic segmentation.
\newblock In \emph{Proceedings of the IEEE/CVF Conference on Computer Vision and Pattern Recognition}, pp.\  12537--12546, 2021.

\bibitem[Zhu et~al.(2023)Zhu, Chen, Yin, See, and Liu]{zhu2023continual}
Zhu, L., Chen, T., Yin, J., See, S., and Liu, J.
\newblock Continual semantic segmentation with automatic memory sample selection.
\newblock In \emph{Proceedings of the IEEE/CVF Conference on Computer Vision and Pattern Recognition}, pp.\  3082--3092, 2023.

\bibitem[Zhu et~al.(2024)Zhu, Chen, Yin, See, and Liu]{zhu2024addressing}
Zhu, L., Chen, T., Yin, J., See, S., and Liu, J.
\newblock Addressing background context bias in few-shot segmentation through iterative modulation.
\newblock In \emph{Proceedings of the IEEE/CVF Conference on Computer Vision and Pattern Recognition}, pp.\  3370--3379, 2024.

\bibitem[Zou et~al.(2024)Zou, Yang, Zhang, Li, Li, Wang, Wang, Gao, and Lee]{zou2024segment}
Zou, X., Yang, J., Zhang, H., Li, F., Li, L., Wang, J., Wang, L., Gao, J., and Lee, Y.~J.
\newblock Segment everything everywhere all at once.
\newblock \emph{Advances in Neural Information Processing Systems}, 36, 2024.

\end{thebibliography}
\bibliographystyle{icml2025}

\newpage
\appendix
\onecolumn
\clearpage
\setcounter{page}{1}
\setcounter{table}{0}
\setcounter{figure}{0}

\section{Details of the KOSCamo+ Dataset and Data Processing Protocols in the Experiment}
There is no existing dataset that have the sketch-image pair in camouflaged object detection. Therefore, we build a new dataset called KOSCamo+ in this study. The dataset will be released under CC BY-NC-ND license. 
In dataset acquisision, we designed a custom software on iPad for this task, allowing volunteers to sketch with Apple Pencils directly on the images without any ground truth (GT) mask reference. To keep the sketches efficient and practical, we set a drawing limit of 30 seconds and timed each session, with the average completion time being 21 seconds. After the sketch is obtained, the result is cross-checked by anthor annotator for quality control. 

Figure \ref{sup_fig1} are more visualized samples in our dataset. The first two row shows the original image and GT mask from the existing camouflaged object detection dataset. The fourth row is the ``scribble" data (which does not belong to the KOSCamo+ dataset, but we also draw scribbles for test set samples for performance comparison), which shows significant differences compared to sketches. Sketches are more precise, showing a rough contour of the object (but not quite accurate), while scribble only depicts the internal area, which is more abstract. The scribble need to be draw strictly inside the object, while our sketches does not have such the restriction. In our main experiment, the sketch input is from the proposed KOSCamo+ dataset, the box input is derived from the bounding rectangle of the Sketch, the Point is the center coordinate of the box to make sure it is within the object. In comparing UniREF++ and RefCod, since these method can takes images as the input, we take scribble, sketch, point, in images and feed into the network.

\begin{figure}[!htbp]
    \centering
    \includegraphics[width=0.8\linewidth]{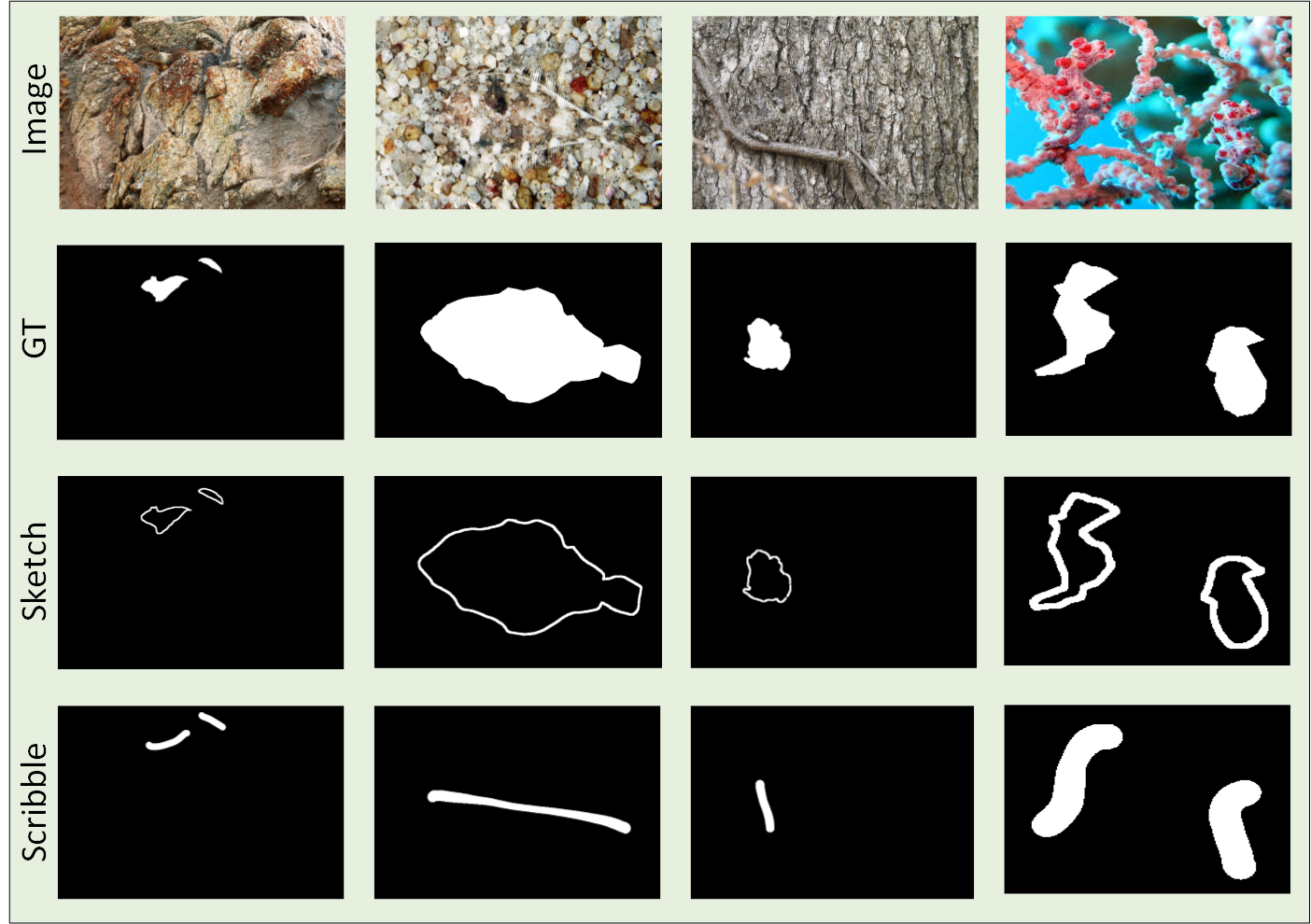}
    \caption{Comparison of freehand sketch and scribble.}
    \label{sup_fig1}
\end{figure}
\vskip -0.5in

\section{More Implementation Details}
In our main experiment, it is worth noting that although we borrowed the image encoder, prompt encoder and mask decoder from Segment Anything (SAM), the information fusion mechanism and the image encoder are different compared to SAM for better interaction between the features of the image and the sketch. The illustration of the differences is shown in Figure \ref{sup_sam}. Specifically, the input to our Prompt Encoder is the features extracted by ResNet18. The Image Encoder injects high-frequency components of the image, extracted using Fourier transforms through different Adapters, and fuses these extracted features. The adapter consists of two MLPs and an activate func
tion within two MLPs. \text{MLP} 32 linear layers and $\text{MLP}_{up}$ is one linear layer that maps the output from GELU activation. The number of adapters are the number of inputs of the transformer layer. We use ViT-H version of SAM image encoder. The the prompt decoder and the mask decoder are initialized with SAM's pre-trained weight and is set to be trainable during the entire training process. The adapter is first pre-trained using the camouflaged dataset (the same dataset in our main experiment) with GT supervision. The SAM image encoder is fixed in the above-mentioned pre-training stage. AdamW optimizer is used with the initial learning rate of to 2e-4; cosine decay is applied to the learning rate in the pre-training. The pre-training takes 20 epochs. After the pre-training the adapter, the weight is initialized and both the adapter and the SAM image encoder is set trainable in the entire training process. AdamW optimizer is used with a batch size of 32. The total losses contain the mask binary
cross-entropy loss and DICE loss. The loss coefficients in Eq.~\ref{eq:total_loss} are set as  $\lambda_{mask}=2.0$, $\lambda_{dice}=5.0$, $\lambda_{adaptive}=5 \times 10^{-4}$, $\lambda_{boundary}=1.0$, respectively. The overall loss function is:  

\vspace{-3mm}
\begin{align}
    \mathcal{L} = \;  
    &\lambda_{mask} \cdot \mathcal{L}_{\mathit{mask}} +
    \lambda_{dice} \cdot \mathcal{L}_{\mathit{dice}} +  \lambda_{adaptive} \cdot \mathcal{L}_{\mathit{adaptive}} + \lambda_{boundary} \cdot \mathcal{L}_{\mathit{boundary}}\\
\label{eq:total_loss}
\end{align}
\vspace{-1mm}

\vskip -0.1in
\begin{figure}[!htbp]
    \centering
    \includegraphics[width=0.6\linewidth]{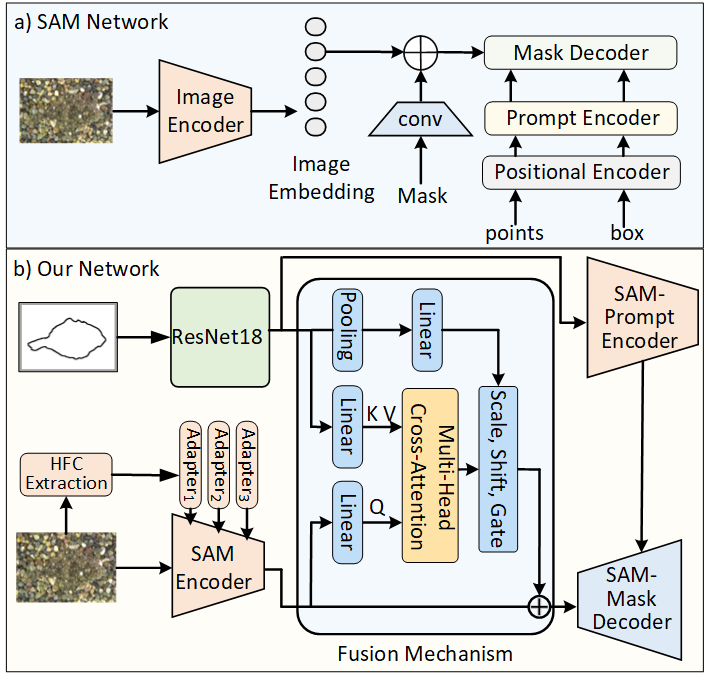}
    \caption{The difference between SAM and our network structure.}
    \label{sup_sam}
\end{figure} 
\vskip -0.3in

\section{More Results for Spatial specificity and Input Error-Tolerance Flexibility for sketch input}

Our sketch input, serving as a prompt similar to the box, scribble, or point inputs in interactive segmentation, focuses on improving the network's performance in camouflaged object detection. Unlike methods like sketch-a-segmenter, our approach incorporates \textbf{spatial specificity} and is specifically designed for camouflaged object detection, meaning the user must \textbf{specify the correct location for the sketch}. Drawing the sketch in the wrong location will result in the network failing to segment the camouflaged object. The second and third columns in the figure \ref{sup_fig2} demonstrate this. The second column shows that when no prompt is given (input is empty), the network's output is also empty, which aligns with our expectations. The third column shows that when the prompt is input in a different location, the network's output does not correspond to the desired camouflaged object, and in most cases, the result is empty, as no camouflaged object is found.

The fourth and fifth columns in the Figure \ref{sup_fig2} further validate the ``the better the drawing, the better the result." When only a partial area or inaccurate contours are drawn, the segmentation of the camouflaged object is also partial. These results demonstrate that more precise and complete sketches lead to better segmentation outcomes, highlighting the importance of the sketch's quality in guiding the network to accurately detect and segment the camouflaged object. 

\begin{figure}[!htbp]
    \centering
    \includegraphics[width=0.5\linewidth]{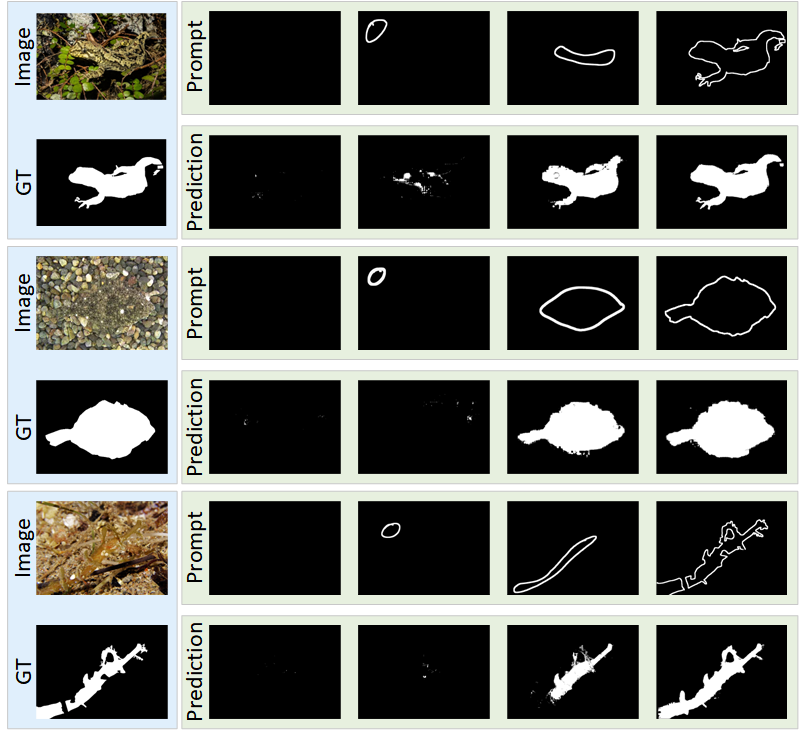}
    \caption{Sketching the “wrong” region. The experiment shows that our network has successfully used the sketch information as
the key reference information.}
    \label{sup_fig2}
\end{figure}

\section{More Result for Ablation Study}
\subsection{The Sensitivity Analysis for $\theta_1$ and $\theta_2$ in Boundary Refinement}
The combination of $\theta_1$ and $\theta_2$ in Boundary Refinement has a certain impact on network performance. We conducted analysis on three datasets to investigate the best value, as shown in Figure \ref{sup-fig4}. We choose $\theta_1$ and $\theta_2$ to be 3, and 3 based on the experimental result.


\begin{figure}[!htbp]
    \centering
    \includegraphics[width=0.9\linewidth]{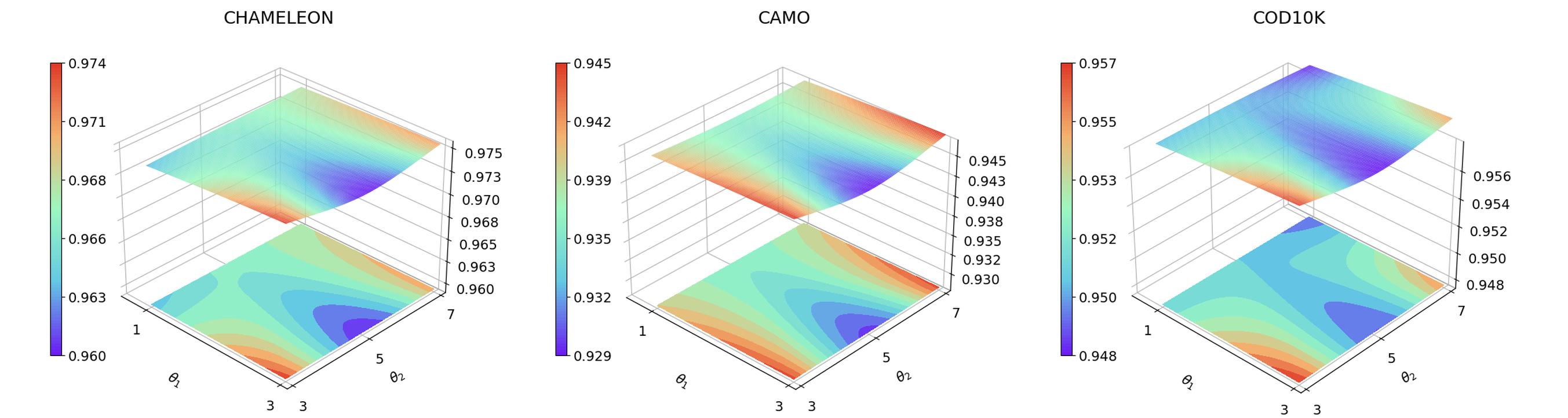}
    \caption{Sensitivity Anallysis of $E_m$ on the CHAMELEON, CAMO, COD10K datasets.}
    \label{sup-fig4}
  \end{figure}

\subsection{Domain Specific Information Injection.}
In Table \ref{sup-ablation study}, we show that by incorporating the patch embedding and high-frequency component as the visual prompt into the adapter, the network performance can be boosted in all dataset. ``inj info" denotes using the backbone network with the adapter in the SAM encoder  (as shown in Figure 2 in the supplementary material), while ``baseline" means using the pre-trained SAM encoder and w/o the adapter while keeping other configurations the same.

\begin{table*}[!htbp]
\caption{Ablation study on the domain-specific info (Inj. Info.).}
\label{sup-ablation study}
\vskip 0.1in
\centering
\resizebox{0.45\linewidth}{!}{
\begin{tabular}{ll|llll}
\hline
\multicolumn{2}{c|}{Method}                  & $S_m \uparrow$    & $E_m \uparrow$    & $F_\beta^\omega \uparrow$ & MAE$\downarrow$   \\ \hline
\multicolumn{1}{l|}{Baseline} & Inj.Info.    & \multicolumn{4}{c}{CHAMELEON (76 images)}                                             \\ \hline
$\checkmark$                  & $\times$     & 0.906 & 0.966 & 0.881         & 0.023 \\
$\checkmark$                  & $\checkmark$ & 0.903 & 0.967 & 0.879         & 0.024 \\ \hline
\multicolumn{2}{l|}{}                        & \multicolumn{4}{c}{CAMO (250 images)}                                                 \\ \hline
$\checkmark$                  & $\times$     & 0.856 & 0.942 & 0.833         & 0.049 \\
$\checkmark$                  & $\checkmark$ & 0.861 & 0.943 & 0.840         & 0.047 \\ \hline
\multicolumn{2}{l|}{}                        & \multicolumn{4}{c}{COD10K (2026 images)}                                              \\ \hline
$\checkmark$                  & $\times$     & 0.867 & 0.950 & 0.807         & 0.023 \\
$\checkmark$                  & $\checkmark$ & 0.872 & 0.954 & 0.817         & 0.021 \\ \hline
\end{tabular}}
\vskip -0.1in
\end{table*}

\subsection{Adaptive Focal Loss, Boundary Refinement, and Sketch Augmentation}
Adaptive Focal Loss, Boundary Refinement, and Sketch Augmentation all contribute to the enhancement of network performance. The full comparison is shown in Table \ref{table.ablation supple}. Adaptive, Boundary, and SA correspond to Adaptive Focal Loss, Boundary Refinement, and Sketch Augmentation, respectively in the table. All experiment shares the same training protocol with the same network backbone (as shown in Figure. 2 in the supplementary material).

\begin{table}[!htbp]
\caption{The effectiveness of Adaptive Focal Loss, Boundary Refinement and Sketch Augmentation.}
\centering
\vskip 0.1in

\resizebox{0.5\linewidth}{!}{
\begin{tabular}{lll|llll}
\hline
\multicolumn{3}{c|}{Method}                                                              & \multicolumn{1}{c}{$S_m \uparrow$} & \multicolumn{1}{c}{$E_{m} \uparrow$} & \multicolumn{1}{c}{$F_\beta^w \uparrow$} & \multicolumn{1}{c}{MAE$ \downarrow$} 
\\ \hline

\multicolumn{1}{c|}{Adaptive} & \multicolumn{1}{c|}{Boundary} & \multicolumn{1}{c|}{SA}    & \multicolumn{4}{c}{CHAMELEON (76 images)}                                                                                                                                  \\ \hline
$\times$                      & $\checkmark$                  & $\checkmark$             & 0.903                  & 0.964                    & 0.878                        & 0.023                    \\
$\checkmark$                  & $\times$                      & $\checkmark$             & 0.904                  & 0.967                    & 0.883                        & 0.022                    \\
$\checkmark$                  & $\checkmark$                  & $\times$                 & 0.904                  & 0.962                    & 0.883                        & 0.023                    \\
$\checkmark$                  & $\checkmark$                  & $\checkmark$             & 0.908                  & 0.972                    & 0.887                        & 0.022  \\
\hline

\multicolumn{3}{c|}{}                                                                    & \multicolumn{4}{c}{CAMO (250 images)}                                                                                                                                    \\ \hline
$\times$                      & $\checkmark$                  & $\checkmark$             & 0.859                  & 0.941                    & 0.839                        & 0.049                    \\
$\checkmark$                  & $\times$                      & $\checkmark$             & 0.850                  & 0.928                    & 0.826                        & 0.049                    \\
$\checkmark$                  & $\checkmark$                  & $\times$                 & 0.862                  & 0.943                    & 0.839                        & 0.045                    \\
$\checkmark$                  & $\checkmark$                  & $\checkmark$             & 0.861                  & 0.945                    & 0.842                        & 0.048                    \\ \hline

\multicolumn{3}{c|}{}                                                                    & \multicolumn{4}{c}{COD10K (2026 images)}                                                                                                                                  \\ \hline
$\times$                      & $\checkmark$                  & $\checkmark$             & 0.874\scriptsize                  & 0.955\scriptsize                 & 0.824\scriptsize                     & 0.020\scriptsize                 \\
$\checkmark$                  & $\times$                      & $\checkmark$             & 0.868\scriptsize                & 0.948\scriptsize                  & 0.813\scriptsize                      & 0.022\scriptsize                 \\
$\checkmark$                  & $\checkmark$                  & $\times$                 & 0.875\scriptsize                & 0.952\scriptsize                 & 0.825\scriptsize                      & 0.021\scriptsize                   \\
$\checkmark$                  & $\checkmark$                  & $\checkmark$             & 0.875\scriptsize               & 0.956\scriptsize                 & 0.825\scriptsize                      & 0.021\scriptsize \\
\hline
\end{tabular}}

\label{table.ablation supple}
\vskip -0.1in
\end{table}

\subsection{Sensitivity analysis of $\theta$ in Adaptive Focal Loss}
To further verify the impact of $\theta$ in Adaptive Focal Loss, we have added an analysis of $S_m$ and $F$ when $\theta$ takes different values, as shown in Figure \ref{sup_fig5}.
It can be observed from the figure that the best performance is achieved when $\theta$ is set to 0.5.

\begin{figure}[!htbp]
    \centering
    \includegraphics[width=\linewidth]{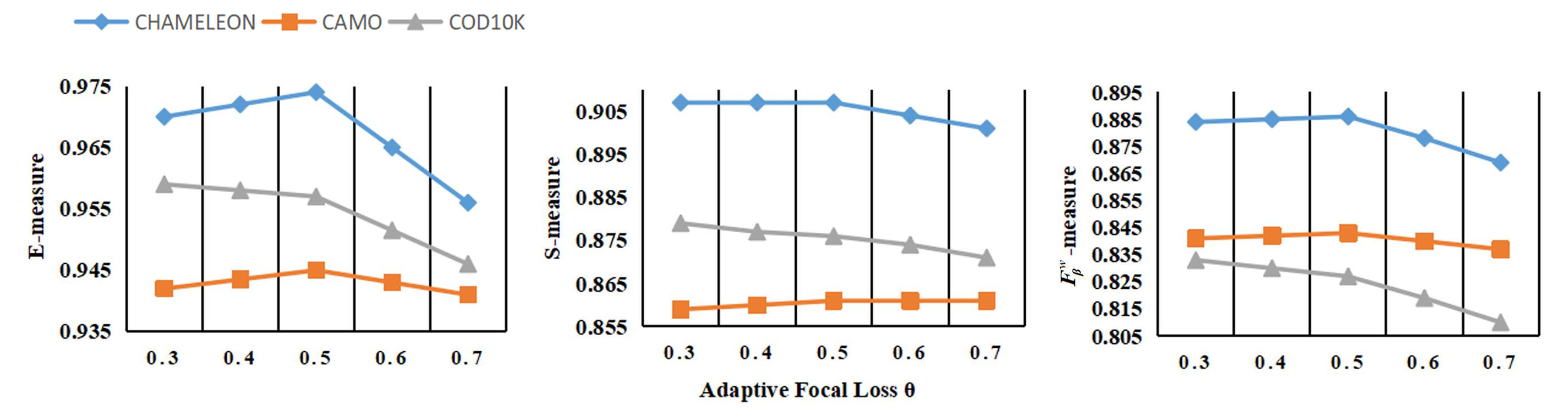}
    \caption{The analysis of $E_m$, $F_{\beta}^w$, MAE metrics on the COD10k dataset.}
    \label{sup_fig5}
\end{figure}

\subsection{Sensitivity analysis for hyperparameter $n$}

Sketch Augmentation divides the image into multiple patches, and we then investigate how the number of these patches $n$ affect the performance. The experimental result is in Table \ref{table.sensitivity supple}, in which we only change the number $n$ while keeping other configurations the same as our final model and our final training protocol. 

\vskip -0.2in

\begin{table}[!htbp]
\caption{Sensitivity Analysis on $n$ in SA on CHAMELEON, CAMO, and COD10K.}
\centering
\vskip 0.1in
\resizebox{0.75\linewidth}{!}{
\begin{tabular}{l|llll|llll|llll}
\hline
\multirow{2}{*}{$n$} & \multicolumn{4}{c|}{CHAMELEON (76 images)} & \multicolumn{4}{c|}{CAMO (250 images)} & \multicolumn{4}{c}{COD10K (2026 images)} \\ \cline{2-13} 
                     & $S_{m}\uparrow$ & $E_{\text{m}} \uparrow$ & $F_{\beta}^{w} \uparrow$ & MAE$\downarrow$ & $S_{m}\uparrow$ & $E_{\text{m}} \uparrow$ & $F_{\beta}^{w} \uparrow$ & MAE$\downarrow$ & $S_{m}\uparrow$ & $E_{\text{m}} \uparrow$ & $F_{\beta}^{w} \uparrow$ & MAE$\downarrow$ \\ \hline
36                   & 0.906           & 0.962                   & 0.887                    & 0.023           & 0.845           & 0.933                   & 0.822                    & 0.052           & 0.859           & 0.945                   & 0.796                    & 0.024           \\
64                   & \textbf{0.908}  & \textbf{0.972}          & \textbf{0.887}           & \textbf{0.022}  & \textbf{0.861}  & \textbf{0.945}          & \textbf{0.842}           & \textbf{0.048}  & \textbf{0.875}  & \textbf{0.956}          & \textbf{0.825}           & \textbf{0.021}  \\
81                   & 0.901           & 0.962                   & 0.887                    & 0.024           & 0.845           & 0.928                   & 0.818                    & 0.052           & 0.860           & 0.944                   & 0.799                    & 0.024           \\ \hline
\end{tabular}}

\label{table.sensitivity supple}
\end{table}

\subsection{Ablation for Sketch Augmentation and ``the better the sketch, the better the outcome"}

We first demonstrate the impact of the sketch augmentation module on the results. As shown in Table. \ref{sup-sa}, ``Augmented" refers to the group where we introduced sketch augmentation. This augmentation is based on edge maps generated by the Canny operator applied to the GT Mask, which enhances the contours by slightly deviating from the GT edges. Specifically, the magnitute is controlled by displacement $\Delta = {\rm floor}\left\lfloor\frac{row}{C}\right\rfloor \times K$. We used K=8 for this enhancement. The ``-" group represents the baseline, where no sketch augmentation is applied, and training is conducted directly on the edge map generated from the GT Mask. In our experiments, all other components and the training process remain consistent, with the only variation being the input sketch.

The experimental results confirm that adding sketch augmentation improves performance, which aligns with our intuition as it enhances the network's robustness to sketches of varying accuracy and styles. Furthermore, our proposed sketch augmentation enables us to obtain sketches with different levels of abstraction (deviation from the true value), denoted as ``weak augmentation" with K=8 and ``strong augmentation," with K=20. The augmentation is imposed to the sketches in KOSCamo+ dataset. From the experimental results, we can observe that applying strong augmentation to the test set leads to poorer network predictions. This outcome is intuitive: the less accurate the sketch, the worse the result.


\begin{table}[!htbp]
\caption{Different Sketch Enhancement/Perturbations in training/testing data affect the performance.}
\label{sup-sa}
\vskip 0.1in
\centering
\renewcommand{\arraystretch}{1.2}
\setlength{\tabcolsep}{4pt}
\begin{tabular}{l|l|cccc|cccc|cccc}
\hline
\multirow{2}{*}{Training data} & \multirow{2}{*}{Test Data} & \multicolumn{4}{c|}{CHAMELEON (76 images)}                                                                                                            & \multicolumn{4}{c|}{CAMO (250 images)}                                                                                                                & \multicolumn{4}{c}{COD10K (2026 images)}                                                                                                             \\ \cline{3-14} 
                               &                            & \multicolumn{1}{c|}{$S_{m}\uparrow$} & \multicolumn{1}{c|}{$E_{\text{m}} \uparrow$} & \multicolumn{1}{c|}{$F_{\beta}^{w} \uparrow$} & MAE$\downarrow$ & \multicolumn{1}{c|}{$S_{m}\uparrow$} & \multicolumn{1}{c|}{$E_{\text{m}} \uparrow$} & \multicolumn{1}{c|}{$F_{\beta}^{w} \uparrow$} & MAE$\downarrow$ & \multicolumn{1}{c|}{$S_{m}\uparrow$} & \multicolumn{1}{c|}{$E_{\text{m}} \uparrow$} & \multicolumn{1}{c|}{$F_{\beta}^{w} \uparrow$} & MAE$\downarrow$ \\ \hline
Augmented                      & -                          & \multicolumn{1}{c|}{\textbf{0.907}}  & \multicolumn{1}{c|}{\textbf{0.974}}          & \multicolumn{1}{c|}{\textbf{0.886}}           & \textbf{0.022}  & \multicolumn{1}{c|}{\textbf{0.861}}  & \multicolumn{1}{c|}{\textbf{0.945}}          & \multicolumn{1}{c|}{\textbf{0.843}}           & \textbf{0.048}           & \multicolumn{1}{c|}{\textbf{0.876}}  & \multicolumn{1}{c|}{\textbf{0.957}}          & \multicolumn{1}{c|}{\textbf{0.827}}           & \textbf{0.021}  \\
Augmented                      & Weak                       & \multicolumn{1}{c|}{0.905}           & \multicolumn{1}{c|}{0.968}                   & \multicolumn{1}{c|}{0.883}                    & 0.023           & \multicolumn{1}{c|}{0.847}           & \multicolumn{1}{c|}{0.925}                   & \multicolumn{1}{c|}{0.822}                    & 0.059           & \multicolumn{1}{c|}{0.872}           & \multicolumn{1}{c|}{0.952}                   & \multicolumn{1}{c|}{0.818}                    & \textbf{0.021}  \\
Augmented    w                  & Strong                     & \multicolumn{1}{c|}{0.905}           & \multicolumn{1}{c|}{0.969}                   & \multicolumn{1}{c|}{0.880}                    & 0.023           & \multicolumn{1}{c|}{0.844}           & \multicolumn{1}{c|}{0.923}                   & \multicolumn{1}{c|}{0.817}                    & 0.061           & \multicolumn{1}{c|}{0.871}           & \multicolumn{1}{c|}{0.950}                   & \multicolumn{1}{c|}{0.815}                    & 0.022           \\
-                              & -                          & \multicolumn{1}{c|}{0.904}           & \multicolumn{1}{c|}{0.962}                   & \multicolumn{1}{c|}{0.883}                    & 0.023           & \multicolumn{1}{c|}{0.862}           & \multicolumn{1}{c|}{0.943}                   & \multicolumn{1}{c|}{0.839}                    & 0.045 & \multicolumn{1}{c|}{0.875}           & \multicolumn{1}{c|}{0.952}                   & \multicolumn{1}{c|}{0.825}                    & \textbf{0.021}  \\
-                              & Weak                       & \multicolumn{1}{c|}{0.902}           & \multicolumn{1}{c|}{0.963}                   & \multicolumn{1}{c|}{0.875}                    & 0.023           & \multicolumn{1}{c|}{0.848}           & \multicolumn{1}{c|}{0.921}                   & \multicolumn{1}{c|}{0.822}                    & 0.057           & \multicolumn{1}{c|}{0.870}           & \multicolumn{1}{c|}{0.947}                   & \multicolumn{1}{c|}{0.815}                    & 0.022           \\
-                              & Strong                     & \multicolumn{1}{c|}{0.898}           & \multicolumn{1}{c|}{0.956}                   & \multicolumn{1}{c|}{0.868}                    & 0.025           & \multicolumn{1}{c|}{0.844}           & \multicolumn{1}{c|}{0.918}                   & \multicolumn{1}{c|}{0.819}                    & 0.060           & \multicolumn{1}{c|}{0.869}           & \multicolumn{1}{c|}{0.947}                   & \multicolumn{1}{c|}{0.812}                    & 0.023           \\ \hline
\end{tabular}
\vskip -0.1in
\end{table}


\section{Confidence Map Visualization}
When we introduce sketches into the segmentation network, we observe some interesting behavior in challenging regions. For example, in the second column of the Figure \ref{sup-fig:confidence} (baseline), the region to the right of the camouflaged object is effectively segmented by the network, while easier areas, such as the region on the left, exhibit low confidence, almost as if the network has ignored them. Our investigation suggests that this might stem from the limitations of focal loss. By incorporating adaptive focal loss (ours), we were able to segment more areas, albeit with lower confidence. Further research is needed to explore how to improve confidence and enhance the network's ability to segment camouflaged objects more effectively.

\begin{figure}[!htbp]
    \centering
    \includegraphics[width=0.75\linewidth]{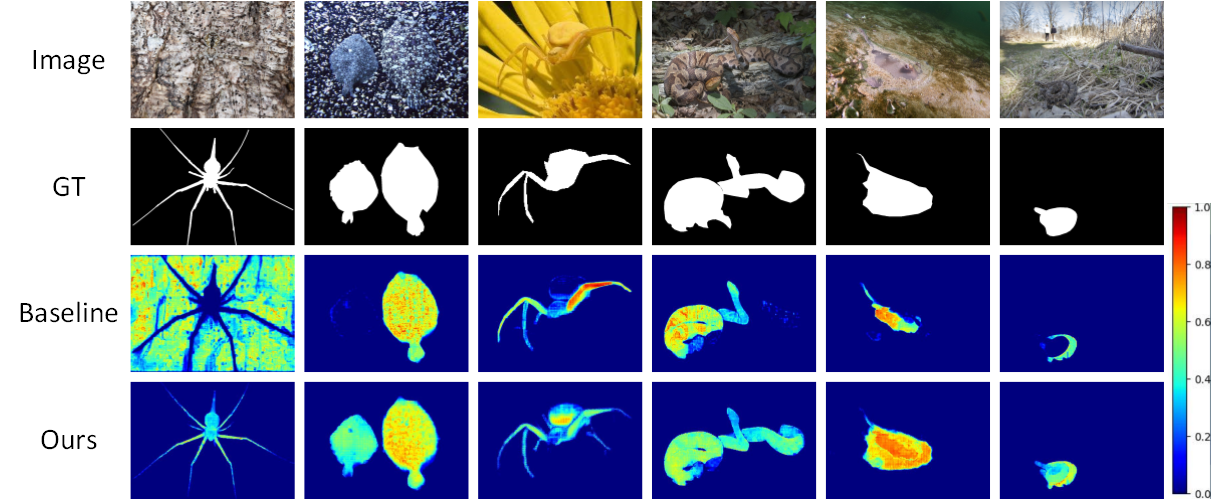}
    \vskip -0.1in
    \caption{The network's confidence heatmap, indicating the regions with the highest confidence.}
    \label{sup-fig:confidence}
\end{figure}

\section{Details About our Applications on Annotation (Sec. 4.3 in the main manuscript)}
We first pre-train our network following the settings mentioned in the manuscript, we then use the pre-trained network to infer segmentation masks in NC4K dataset. The NC4K dataset is a new data set that has not been seen in the pre-training stage. A total of 4,121 numbers of samples from NC4K are used in training. We then use the obtained masks to train two classic camouflaged object detection network, ZoomNet and UGTR, the details of training these two networks are as follows:

\noindent\textbf{Traing ZoomNet.} The encoder utilizes ResNet50 parameters pre-trained on ImageNet, while the remaining parts are randomly initialized. Optimization is performed using SGD with a momentum of 0.9 and a weight decay of 0.0005. The learning rate is initialized to 0.05 and follows a linear warm-up strategy, followed by a linear decay. The model is trained end-to-end on 8 NVIDIA 3090 GPUs for 40 epochs, with a batch size of 8. During both training and inference, the input resolution is set to 384 × 384.

\noindent\textbf{Traing UGTR.} During training, the backbone is initialized with ResNet50 pre-trained on ImageNet, while other components (including the probabilistic module, prototyping transformer, and uncertainty-guided transformer) are initialized randomly. In line with standard practices, data augmentation is applied to each training sample, including random cropping, horizontal flipping, and scaling within the range [0.75, 1.25]. The UGTR model is optimized using Stochastic Gradient Descent (SGD) with a ‘poly’ learning rate schedule. The base learning rate is empirically set to $10^{-7}$, with the power parameter set to 0.9.

After the UGTR and Zoomnet has been trained, we test the network performance in classical benchmarks in camouflaged object detection. The full comparison table is as follows: 

\vskip -0.3in
\begin{table}[!htbp]
\caption{Annotations generated by our method are used as pseudo
labels for ZoomNet and UGTR model training.}

\vskip 0.1in
\centering
\resizebox{0.95\textwidth}{!}{ 
\begin{tabular}{l|llll|llll|llll}
\hline

\multirow{2}{*}{Trained with label} & \multicolumn{4}{c|}{CHAMELEON (76 images)} & \multicolumn{4}{c|}{CAMO (250 images)} & \multicolumn{4}{c}{COD10K (2026 images)} \\ \cline{2-13} 
                              & $S_m \uparrow$   & $E_m \uparrow$  & $F_\beta^\omega \uparrow$ &  MAE$\downarrow$     
                              & $S_m \uparrow$   & $E_m \uparrow$  & $F_\beta^\omega \uparrow$ &  MAE$\downarrow$     
                              & $S_m \uparrow$   & $E_m \uparrow$  & $F_\beta^\omega \uparrow$ &  MAE$\downarrow$ \\ \hline
ZoomNet (Pixel-by-Pixel Annotated)  & 0.845  & 0.919  & 0.758  & 0.034  & 0.769  & 0.836  & 0.668  & 0.083  & 0.818  & 0.884  & 0.688  & 0.034  \\
\textbf{ZoomNet (Our Predicted Label)} & 0.834  & 0.925  & 0.750  & 0.038  & 0.754  & 0.834  & 0.653  & 0.089  & 0.805  & 0.878  & 0.673  & 0.036  \\ \hline
UGTR (Pixel-by-Pixel Annotated)      & 0.872  & 0.944  & 0.761  & 0.039  & 0.761  & 0.834  & 0.630  & 0.096  & 0.792  & 0.864  & 0.601  & 0.046  \\
\textbf{UGTR (Our Predicted Label)}  & 0.877  & 0.946  & 0.760  & 0.037  & 0.766  & 0.825  & 0.634  & 0.097  & 0.799  & 0.871  & 0.609  & 0.044  \\ \hline
\end{tabular}
} 
\label{sup-Pseudo-label}
\vskip -0.1in
\end{table}

 \vskip -0.25in
\begin{figure}[!htbp]
    \centering
    \includegraphics[width=0.55\linewidth]{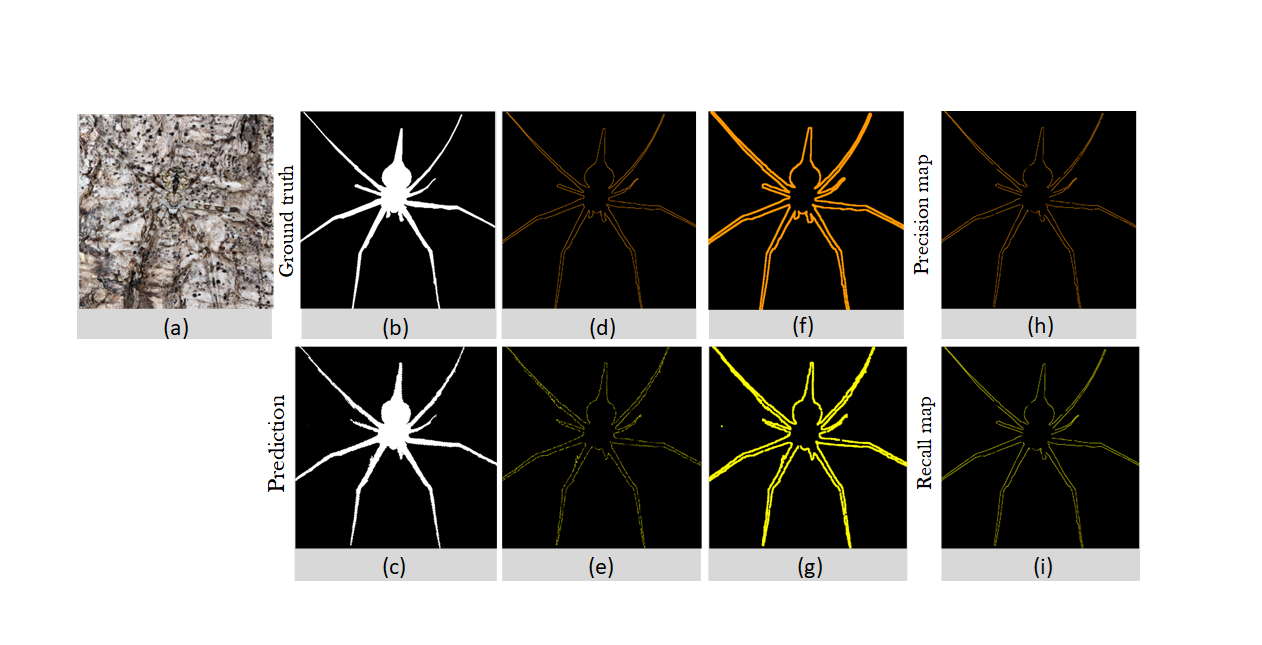}
    \vskip -0.15in
    \caption{Visualization of the implementation process of Boundary Refinement. (a) original image; (b) GT; (c) predicted segment (pred); (d) boundary of GT; (e) boundary of pred; (f) expanded boundary of GT; (g) expanded boundary of pred; (h) pixel-wise multiplication of masks (d) and (g); (i) pixel-wise multiplication of masks (e) and (f).}
    \label{sup-bloss}
\end{figure}

\section{Details About Boundary Refinement (Sec. 3.5 in the main manuscript)}

To demonstrate the functionality of Boundary Refinement, we present the working principle of this loss in Figure \ref{sup-bloss}. In the top row (see (b), (d), and (f)), we show the ground truth, where (b) is the binary mask of the true segmentation, (d) is the binary mask of the true boundary obtained after applying Eq.\ref{equation.3} and Eq.\ref{equation.4}. Then, we compute the expanded boundary (f) using pooling operations. The same pipeline is applied to the predicted mask (with color intensity representing the probability that a pixel belongs to the foreground). Afterward, we obtain the Precision map, which is the pixel-wise multiplication of the maps corresponding to (d) and (g), and the Recall map, which is the pixel-wise multiplication of the maps (e) and (f). Precision and Recall are the pixel-wise sum of these maps, normalized accordingly.

\section{Annotation Time Comparison}

{\begin{figure}[!ht]
    \centering
    \includegraphics[width=0.65\linewidth]{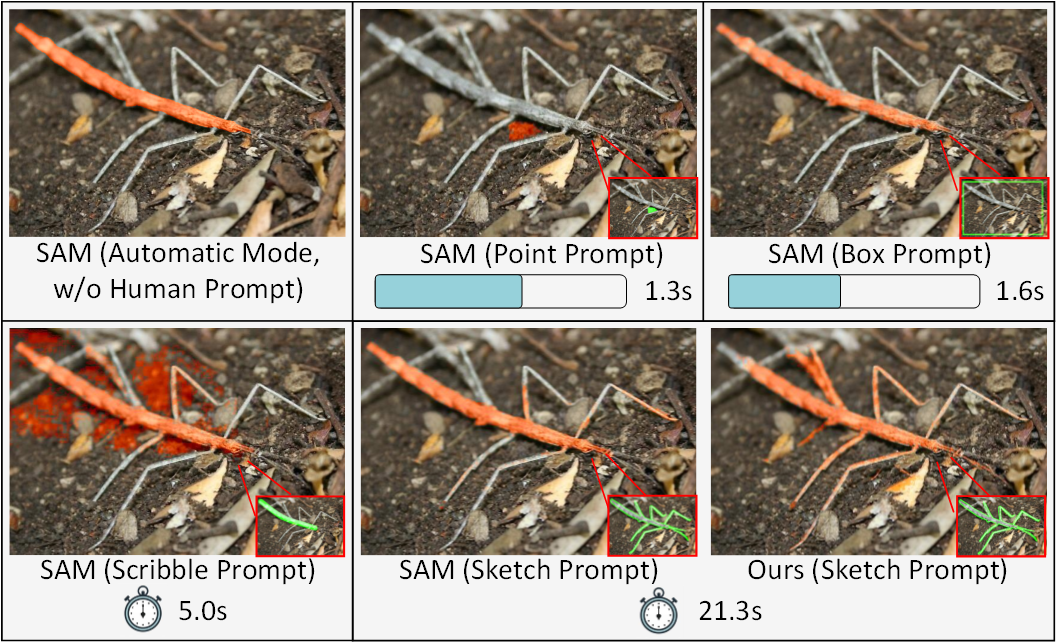}
    \vskip -0.1in
    \caption{Different types of annotation camouflaged object detection task. The mask annotation takes about 60 minutes for each image. The point takes just 2 seconds, the box takes about 3 seconds, the scribble takes 5 seconds and sketch is drawn by different people only takes between 10-30s}
    \vskip -0.2in
    \label{sup-sketch-time}
  \end{figure}
  \vskip -0.1in

\end{document}